\documentclass{article}
\usepackage{PRIMEarxiv}
\usepackage[utf8]{inputenc} 
\usepackage[T1]{fontenc}    
\usepackage{hyperref}       
\usepackage{url}            
\usepackage{booktabs}       
\usepackage{amsfonts}       
\usepackage{nicefrac}       
\usepackage{microtype}      
\usepackage{lipsum}
\usepackage{fancyhdr}       
\usepackage{graphicx}       
\graphicspath{{media/}}     

\usepackage{natbib}
\usepackage{amsmath} 
\usepackage{amssymb}
\usepackage{xcolor}
\usepackage{subcaption}
\usepackage{tabularx}
\usepackage{multirow}
\usepackage{algorithm}
\usepackage{algpseudocode}
\usepackage{comment}
\usepackage{float}
\usepackage{booktabs}
\usepackage{bm}  
\usepackage{placeins}

\title{Meta-Learned Basis Adaptation for Parametric Linear PDEs}

\author{
  Vikas Dwivedi\thanks{Corresponding Author} \\
  CREATIS Biomedical Imaging Laboratory \\
  INSA, CNRS UMR 5220, Inserm, Universit´e Lyon 1 \\
  Lyon 69621, France\\
  \texttt{vikas.dwivedi@creatis.insa-lyon.fr} \\
     \And
  Monica Sigovan \\
  CREATIS Biomedical Imaging Laboratory \\
  INSA, CNRS UMR 5220, Inserm, Universit´e Lyon 1 \\
  Lyon 69621, France\\
  \texttt{monica.sigovan@insa-lyon.fr} \\
    \And
  Bruno Sixou \\
  CREATIS Biomedical Imaging Laboratory \\
  INSA, CNRS UMR 5220, Inserm, Universit´e Lyon 1 \\
  Lyon 69621, France\\
  \texttt{bruno.sixou@insa-lyon.fr} \\
}

\begin{document}
\maketitle

\begin{abstract}
	We propose a hybrid physics-informed framework for solving families of parametric linear partial differential equations (PDEs) by combining a meta-learned predictor with a least-squares corrector. The predictor, termed \textbf{KAPI} (Kernel-Adaptive Physics-Informed meta-learner), is a shallow task-conditioned model that maps query coordinates and PDE parameters to solution values while internally generating an interpretable, task-adaptive Gaussian basis geometry. A lightweight meta-network maps PDE parameters to basis centers, widths, and activity patterns, thereby learning how the approximation space should adapt across the parametric family. This predictor-generated geometry is transferred to a second-stage corrector, which augments it with a background basis and computes the final solution through a one-shot physics-informed Extreme Learning Machine (PIELM)-style least-squares solve. We evaluate the method on four linear PDE families spanning diffusion, transport, mixed advection--diffusion, and variable-speed transport. Across these cases, the predictor captures meaningful physics through localized and transport-aligned basis placement, while the corrector further improves accuracy, often by one or more orders of magnitude. Comparisons with parametric PINNs, physics-informed DeepONet, and uniform-grid PIELM correctors highlight the value of predictor-guided basis adaptation as an interpretable and efficient strategy for parametric PDE solving.
\end{abstract}


\section{Introduction}

Physics-informed machine learning has emerged as a powerful paradigm for solving partial differential equations (PDEs) by combining neural function approximation with the governing equations in the training objective \cite{RAISSI2019686}. Among the most widely studied formulations are Physics-Informed Neural Networks (PINNs), which offer a flexible and general framework for forward and inverse problems. However, standard PINNs are often expensive to train, can struggle in stiff or transport-dominated regimes, and typically require solving a new optimization problem for each new PDE instance \cite{krishnapriyan2021characterizing,WANG2022110768,WANG_2021}. This limitation becomes especially restrictive in \emph{parametric} settings, where one seeks repeated solves across a family of related PDEs \cite{PENWARDEN2023111912,PSAROS2022111121}.

A parallel line of work has explored \emph{shallow} physics-informed models based on structured basis expansions. Physics-Informed Extreme Learning Machines (PIELMs) solve for the output coefficients analytically through least squares while keeping the hidden-layer basis fixed \cite{DWIVEDI_PIELM_2020}. This makes them computationally lightweight, but their performance depends strongly on the chosen hidden basis, and the fixed-input-layer design has been recognized as a key limitation in related shallow collocation formulations \cite{DONG2022111290,CALABRO2021114188}. Recent work by Ramabathiran and Ramachandran introduced SPINN, a sparse and partially interpretable shallow physics-informed architecture in which the hidden layer can be viewed as a mesh-encoding layer with trainable basis centers and widths \cite{ramabathiran2021spinn}. This perspective is important because it injects physics awareness into the \emph{architecture itself}, not only into the loss: centers may be constrained to remain inside the computational domain, widths may adapt to local scales, and the learned node and kernel-width distributions become a direct interpretability diagnostic. In particular, SPINN showed that the geometry of the learned basis can reveal physically important regions such as localized hot spots, shocks, and characteristic-like transport paths.

A closely related development is the curriculum learning-driven PIELM framework of \cite{DWIVEDI2025130924}, which adopted the same SPINN-inspired shallow Gaussian basis philosophy within a PIELM solver. That work showed that, given a sufficiently good hidden basis, shallow least-squares physics solvers can be remarkably effective, even outperforming deeper physics-informed models in several settings. At the same time, it also exposed a central limitation of PIELM-style methods: because the hidden layer remains fixed during training, the basis geometry must still be prescribed heuristically for each PDE instance. Thus, although the hidden basis is interpretable, its initialization remains a largely instance-wise and user-driven design problem.

A third relevant line of work concerns \emph{adaptive} physics-informed solvers. In adaptive PINN frameworks such as residual-based adaptive refinement (RAR), the PDE residual is used to enrich the collocation set, but the network architecture itself typically remains fixed \cite{lu2021deepxde}. More recent shallow adaptive methods extend adaptation beyond collocation to the basis itself, allowing the effective basis distribution to change in response to the PDE residual \cite{DONG2022111290,CALABRO2021114188}. For example, recent kernel-adaptive PI-ELM formulations optimize low-dimensional, physically interpretable kernel-distribution parameters and thereby adapt both collocation and basis allocation \cite{dwivedi2025kerneladaptivepielmsforwardinverse}. However, such approaches remain fundamentally \emph{single-instance}: adaptation is still performed separately for each PDE, and current demonstrations are largely limited to steady problems.

This paper addresses that gap. We ask the following question: \emph{can basis adaptation itself be learned across a family of parametric PDEs, so that each new task receives a task-appropriate approximation space in one shot, without iterative residual refinement at test time?} Our central idea is to treat \emph{basis geometry} as the object to be meta-learned. Concretely, the full predictor is a shallow task-conditioned model, while a lightweight internal meta-network maps the low-dimensional PDE parameter vector to task-dependent basis centers, widths, and activity patterns. In this view, the predictor acts as an \emph{amortized kernel-adaptation module}: rather than optimizing kernel geometry separately for each PDE instance, it learns a shared map from task parameters to task-appropriate basis geometry across the entire family.

Building on this idea, we develop a hybrid predictor-corrector framework. The full predictor, which we call \emph{KAPI} (Kernel-Adaptive Physics-Informed meta-learner), is a shallow task-conditioned model whose task dependence is mediated through an internal meta-network that generates basis geometry from PDE parameters. In this sense, KAPI can be interpreted as a \emph{parametric extension of the SPINN philosophy}: it retains the shallow, interpretable, basis-driven structure of SPINN, but lifts it from the single-instance setting to the family setting. The predictor-generated basis geometry is then passed to a physics-informed corrector, which augments it with a background scaffold and computes the final coefficients through a PIELM-style least-squares solve. This coupling replaces heuristic per-instance basis initialization with a predictor-guided, one-shot correction mechanism. Unlike iterative residual-adaptive PINNs or adaptive ELM schemes, which refine one PDE instance at a time, the proposed framework performs \emph{amortized targeted refinement}: after offline meta-training, inference on a new task consists of one predictor pass followed by one least-squares correction.

Our setting is also distinct from standard neural-operator learning. Neural operators such as DeepONet \cite{Lu2021,Goswami2023} and Fourier Neural Operators (FNO) \cite{li2021fourierneuraloperatorparametric} are designed to learn mappings between function spaces. By contrast, our method targets low-dimensional parametric PDE families and learns a map from PDE parameters to an \emph{interpretable approximation space}, followed by a physics-based solve in that space. The goal is therefore not direct operator regression, but task-conditioned basis generation. Nevertheless, we include limited comparisons with representative parametric neural baselines to show that, within the scope of the present problems, the KAPI predictor is already competitive despite using only a shallow single-hidden-layer hypothesis and offering explicit geometric interpretability.

We evaluate the proposed framework on four representative linear PDE families spanning localized elliptic response, constant-speed transport, mixed advection--diffusion, and variable-speed transport. These families were chosen to test basis adaptation under markedly different physical regimes: localized source-driven solutions, transported Gaussian packets, transport--diffusion competition, and curved variable-speed characteristics. Across these cases, the experiments show a consistent pattern. The predictor alone already identifies the important regions of the solution manifold, such as localized gradient hot spots and transport-aligned space-time corridors. The corrector then exploits this geometry to obtain a more accurate final solution, often improving the predictor by one or more orders of magnitude. The accompanying geometry plots provide a direct explanation for this behavior by showing how basis centers, widths, and activity patterns align with the underlying physics.

\paragraph{Main contributions.}
The main contributions of this work are as follows:
\begin{itemize}
	\item We introduce a hybrid physics-informed framework for parametric linear PDEs that combines a meta-learned shallow predictor with a PIELM-style least-squares corrector. The key idea is to meta-learn task-adaptive basis geometry and then solve each new PDE instance in this enriched basis through a one-shot physics-informed correction.
	
	\item We extend the SPINN philosophy of interpretable shallow basis geometry from the single-instance setting to the parametric setting. In the proposed KAPI predictor, an internal lightweight meta-network plays the role of a family-level kernel-adaptation module, generating task-dependent basis centers, widths, and activity patterns directly from PDE parameters.
	
	\item We replace the heuristic hidden-layer initialization required by PIELM-style solvers with predictor-guided basis generation. This yields a simple alternative to iterative residual-adaptive refinement strategies: basis adaptation is amortized across a PDE family and deployed in one shot at inference time after offline meta-training.
	
	\item Through experiments on four PDE families spanning diffusion-dominated, transport-dominated, mixed advection--diffusion, and variable-speed transport regimes, we show that the predictor alone already captures significant physics, while the corrector often improves accuracy by one or more orders of magnitude, including in several extrapolative settings.
	
	\item We provide targeted ablation studies showing that the final gain comes specifically from predictor-guided basis adaptation. In particular, the proposed corrector substantially outperforms uniform-grid PIELM baselines, and the comparison with single-instance PINNs clarifies the advantages and disadvantages between amortized family-level solving and per-instance optimization.
\end{itemize}

\paragraph{Organization of the paper.}
The remainder of the paper is organized as follows. Section~\ref{sec:math_formulation} presents the mathematical formulation of the predictor-corrector framework. Section~\ref{sec:problem_definition} introduces the four parametric PDE families used as test cases. Section~\ref{sec:results} reports the main numerical results, including predictor comparisons with parametric neural baselines, predictor-corrector performance across PDE families, interpretability analyses, and ablation studies. Section~\ref{sec:limitations} discusses the main limitations of the current study. Finally, Section~\ref{sec:conclusion} concludes the paper.
\section{Mathematical Formulation}
\label{sec:math_formulation}
\begin{figure*}[t]
	\centering
	\includegraphics[width=0.85\textwidth]{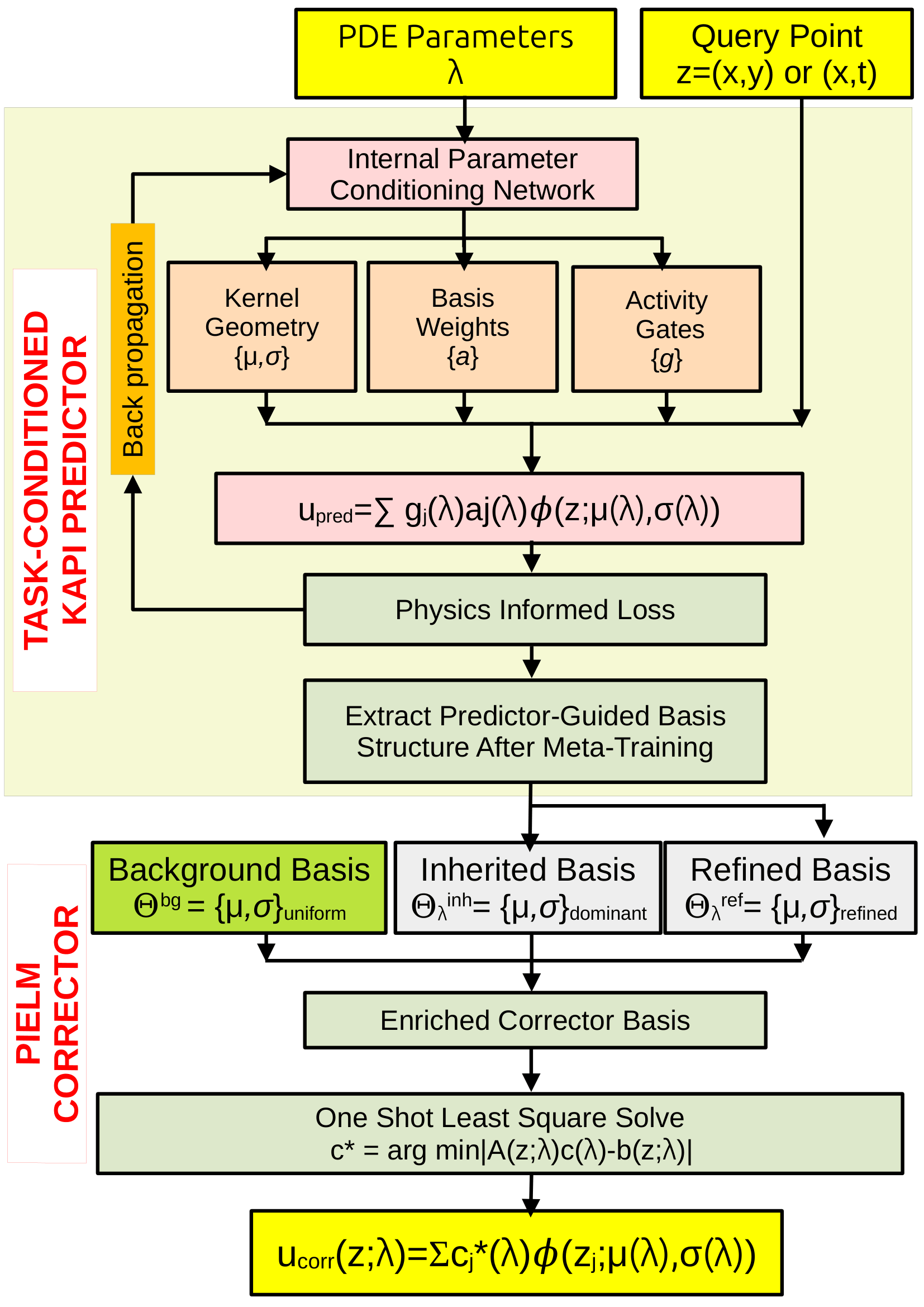}
	\caption{
		Overview of the proposed predictor-corrector framework. The full KAPI predictor is a shallow task-conditioned model that takes PDE parameters $\lambda$ and a query point $\mathbf z$ and produces a coarse prediction $u^{\mathrm{pred}}(\mathbf z;\lambda)$. Its task dependence is mediated through an internal parameter-conditioning network, which generates predictor basis structure including centers, widths, activity patterns, and other task-dependent basis parameters. The corrector then inherits the most informative predictor-guided basis elements, augments them with predictor-informed refinement and a background scaffold, and computes the final coefficients through a one-shot PIELM-style least-squares solve. The resulting corrected solution $u^{\mathrm{corr}}(\mathbf z;\lambda)$ combines amortized task-adaptive basis generation with physics-constrained coefficient recovery.
	}
	\label{fig:opener}
\end{figure*}

Figure~\ref{fig:opener} summarizes the overall predictor-corrector framework. The full KAPI predictor is a shallow task-conditioned model that produces a coarse solution estimate while internally generating a task-adaptive basis structure from the PDE parameters. This predictor-generated basis structure is then transferred to a second-stage corrector, which augments it with refinement and background support and computes the final coefficients through a one-shot least-squares solve. The proposed framework combines two familiar ingredients from physics-informed learning. The predictor is trained in a PINN-like manner, in the sense that its basis geometry is optimized through a physics-informed loss. The corrector is PIELM-like, in the sense that once a hidden basis is fixed, the final output coefficients are computed by a one-shot least-squares solve rather than by iterative backpropagation. Our method differs from both classical settings because the hidden basis is not fixed a priori and is not optimized separately for each PDE instance; instead, it is generated in a task-conditioned manner by the predictor and then reused by the corrector. We now formalize each component in turn.

We consider a family of parametric linear PDEs indexed by a finite-dimensional parameter vector
\[
\lambda \in \Lambda \subset \mathbb{R}^{d_\lambda},
\]
where $d_\lambda$ denotes the dimension of the parameter space. For each parameter value $\lambda$, let the exact solution $u^\star(\cdot;\lambda)$ satisfy
\begin{equation}
	\mathcal{L}_\lambda u(\mathbf{z};\lambda)=f(\mathbf{z};\lambda),
	\qquad
	\mathbf{z}\in\Omega,
	\label{eq:pde_family}
\end{equation}
together with the appropriate boundary conditions
\begin{equation}
	\mathcal{B}_\lambda u(\mathbf{z};\lambda)=g(\mathbf{z};\lambda),
	\qquad
	\mathbf{z}\in\partial\Omega,
	\label{eq:bc_family}
\end{equation}
and, when relevant, initial conditions
\begin{equation}
	\mathcal{I}_\lambda u(\mathbf{z};\lambda)=u_0(\mathbf{z};\lambda).
	\label{eq:ic_family}
\end{equation}

Rather than solving each parameterized PDE instance independently, we assume that parameter values are sampled from a distribution
\[
\lambda \sim \rho(\lambda),
\]
and seek a single solver that exploits shared structure across the family.

\subsection{Meta-Learned Predictor}
\label{subsec:stage1_predictor}

The full predictor in our framework is a task-conditioned shallow model
\[
u_\theta^{\mathrm{pred}}:\Omega\times\Lambda\to\mathbb{R},
\qquad
(\mathbf{z},\lambda)\mapsto u_\theta^{\mathrm{pred}}(\mathbf{z};\lambda),
\]
which maps a physical query point $\mathbf{z}$ and a PDE parameter vector $\lambda$ to a predicted solution value. We refer to this full predictor as \emph{KAPI}, short for \emph{Kernel-Adaptive Physics-Informed} meta-learner.

A key feature of KAPI is that the task dependence is introduced through an internal meta-network
\[
h_\psi:\Lambda\to\mathcal{G},
\]
which takes only the PDE parameter vector $\lambda$ as input and outputs a task-adaptive hidden-basis description. We denote this predictor-generated basis description by
\begin{equation}
	\Gamma_\psi(\lambda)
	=
	\Big\{
	\mu_j(\lambda),\;
	\sigma_j(\lambda),\;
	g_j(\lambda),\;
	a_j(\lambda)
	\Big\}_{j=1}^{M}.
	\label{eq:geometry_set}
\end{equation}
Here:
\begin{itemize}
	\item $\mu_j(\lambda)$ denotes the location of the $j$th basis function,
	\item $\sigma_j(\lambda)$ denotes its width or scale,
	\item $g_j(\lambda)$ denotes an activity gate or importance weight,
	\item $a_j(\lambda)$ denotes a predictor amplitude or coefficient.
\end{itemize}
For steady problems these quantities are purely spatial, whereas for unsteady problems they may encode space-time geometry or dynamic basis evolution. Thus, the meta-network does not directly output the solution field; instead, it outputs the \emph{representation geometry} from which the predictor field is assembled.

Given a query point $\mathbf{z}$, the KAPI predictor evaluates localized basis functions and combines them linearly:
\begin{equation}
	u_\theta^{\mathrm{pred}}(\mathbf{z};\lambda)
	=
	\sum_{j=1}^{M}
	g_j(\lambda)\,a_j(\lambda)\,
	\phi\!\big(\mathbf{z};\mu_j(\lambda),\sigma_j(\lambda)\big),
	\label{eq:predictor_ansatz}
\end{equation}
where $\phi$ denotes a localized basis function. Depending on the PDE family, part of the amplitudes may be shared across tasks, while in other cases they are also task-dependent. In all cases, the predictor remains a structured shallow basis model rather than an unconstrained deep neural field.

Equivalently, KAPI may be viewed as a \emph{family-level kernel-adaptation mechanism}: instead of directly learning a black-box task-to-solution map, it learns how the hidden basis itself should adapt across the parametric task family.

\subsection{Physics-Informed Meta-Training}
\label{subsec:meta_training}

The full KAPI predictor $u_\theta^{\mathrm{pred}}$ is trained over the task distribution $\rho(\lambda)$ using a physics-informed objective. For each task $\lambda$, let
\[
\Omega_\lambda^{\mathrm{int}},\qquad
\Omega_\lambda^{\mathrm{bc}},\qquad
\Omega_\lambda^{\mathrm{ic}}
\]
denote the interior, boundary, and initial collocation sets, respectively. We define the interior residual
\begin{align}
	\mathcal{R}_{\mathrm{int}}(\mathbf{z};\lambda)
	&=
	\mathcal{L}_\lambda u_\theta^{\mathrm{pred}}(\mathbf{z};\lambda)-f(\mathbf{z};\lambda),
	\qquad
	\mathbf{z}\in\Omega_\lambda^{\mathrm{int}},
\end{align}
the boundary residual
\begin{align}
	\mathcal{R}_{\mathrm{bc}}(\mathbf{z};\lambda)
	&=
	\mathcal{B}_\lambda u_\theta^{\mathrm{pred}}(\mathbf{z};\lambda)-g(\mathbf{z};\lambda),
	\qquad
	\mathbf{z}\in\Omega_\lambda^{\mathrm{bc}},
\end{align}
and, when relevant, the initial residual
\begin{align}
	\mathcal{R}_{\mathrm{ic}}(\mathbf{z};\lambda)
	&=
	\mathcal{I}_\lambda u_\theta^{\mathrm{pred}}(\mathbf{z};\lambda)-u_0(\mathbf{z};\lambda),
	\qquad
	\mathbf{z}\in\Omega_\lambda^{\mathrm{ic}}.
\end{align}

The meta-learned predictor parameters are obtained by minimizing
\begin{equation}
	\min_{\theta}\;
	\mathbb{E}_{\lambda\sim\rho}
	\Big[
	\omega_{\mathrm{int}}\,\mathcal{L}_{\mathrm{int}}(\lambda)
	+
	\omega_{\mathrm{bc}}\,\mathcal{L}_{\mathrm{bc}}(\lambda)
	+
	\omega_{\mathrm{ic}}\,\mathcal{L}_{\mathrm{ic}}(\lambda)
	+
	\omega_{\mathrm{reg}}\,\mathcal{L}_{\mathrm{reg}}(\lambda)
	\Big],
	\label{eq:meta_training_objective}
\end{equation}
where
\begin{align}
	\mathcal{L}_{\mathrm{int}}(\lambda)
	&=
	\frac{1}{|\Omega_\lambda^{\mathrm{int}}|}
	\sum_{\mathbf{z}\in\Omega_\lambda^{\mathrm{int}}}
	\big|\mathcal{R}_{\mathrm{int}}(\mathbf{z};\lambda)\big|^2,\\
	\mathcal{L}_{\mathrm{bc}}(\lambda)
	&=
	\frac{1}{|\Omega_\lambda^{\mathrm{bc}}|}
	\sum_{\mathbf{z}\in\Omega_\lambda^{\mathrm{bc}}}
	\big|\mathcal{R}_{\mathrm{bc}}(\mathbf{z};\lambda)\big|^2,\\
	\mathcal{L}_{\mathrm{ic}}(\lambda)
	&=
	\frac{1}{|\Omega_\lambda^{\mathrm{ic}}|}
	\sum_{\mathbf{z}\in\Omega_\lambda^{\mathrm{ic}}}
	\big|\mathcal{R}_{\mathrm{ic}}(\mathbf{z};\lambda)\big|^2.
\end{align}
Here $\mathcal{L}_{\mathrm{reg}}$ is a problem-dependent geometry regularizer acting on the predictor-generated basis structure. The precise form of $\mathcal{L}_{\mathrm{reg}}$ depends on the PDE family. For example, in the Poisson case we use a mild sparsity penalty on the meta-learned gates,
\begin{equation}
	\mathcal{L}_{\mathrm{reg}}^{\mathrm{Pois}}(\lambda)
	=
	\lambda_{\mathrm{sp}}
	\frac{1}{M}\sum_{j=1}^{M}|g_j(\lambda)|,
\end{equation}
which encourages a compact active basis. Since the homogeneous Dirichlet boundary condition is hard-enforced through the trial factor, the practical Poisson loss reduces to a PDE residual term together with this sparsity regularization.

For the dynamic linear advection case, the regularizer additionally controls temporal smoothness of moving centers and prevents pathological width collapse. A representative form is
\begin{equation}
	\mathcal{L}_{\mathrm{reg}}^{\mathrm{Adv}}(\lambda)
	=
	\lambda_{\mathrm{c}}\,\mathcal{L}_{\mathrm{center}}(\lambda)
	+
	\lambda_{\mathrm{w}}\,\mathcal{L}_{\mathrm{width}}(\lambda),
\end{equation}
with
\begin{equation}
	\mathcal{L}_{\mathrm{center}}(\lambda)
	=
	\frac{1}{M}\sum_{j=1}^{M}
	\big|\partial_t \xi_j(t,\lambda)\big|
	+
	\eta\,
	\frac{1}{M}\sum_{j=1}^{M}
	|g_j(\lambda)\alpha_j(t,\lambda)|,
\end{equation}
and
\begin{equation}
	\mathcal{L}_{\mathrm{width}}(\lambda)
	=
	\frac{1}{M}\sum_{j=1}^{M}
	\big(h_j(t,\lambda)-h_j^{\mathrm{target}}(\lambda)\big)^2.
\end{equation}

\subsection{Predictor-Guided Physics-Informed Corrector}
\label{subsec:stage2_corrector}

The KAPI predictor provides a task-adaptive approximation-space geometry, but it is not itself the final solver. Instead, the predictor output is used to build a frozen hidden dictionary for a second-stage physics-informed correction.

To make this precise, recall that the full predictor output for task $\lambda$ is the basis description
\[
\Gamma_\psi(\lambda)
=
\Big\{
\mu_j(\lambda),\;
\sigma_j(\lambda),\;
g_j(\lambda),\;
a_j(\lambda)
\Big\}_{j=1}^{M}.
\]
The corrector does not, in general, reuse all predictor basis functions unchanged. Rather, it transfers from the predictor the task-dependent information that defines where useful hidden basis support should be placed. In practice, this transferred information consists of:
\begin{enumerate}
	\item the most informative predictor basis functions, selected according to problem-dependent activity or amplitude scores,
	\item additional predictor-derived geometric cues, such as sharp local gradients or relatively large PDE residuals.
\end{enumerate}

We denote the inherited predictor-guided basis by
\[
\Theta_\lambda^{\mathrm{inh}}
=
\big\{
\theta_j^{\mathrm{inh}}(\lambda)
\big\}_{j=1}^{M_{\mathrm{inh}}},
\]
where each $\theta_j^{\mathrm{inh}}(\lambda)$ collects the basis-defining parameters of a selected predictor basis function. Depending on the PDE family, this may include spatial or space-time centers, widths, activity information, and dynamic basis descriptors.

We further denote by
\[
\Theta_\lambda^{\mathrm{ref}}
=
\big\{
\theta_j^{\mathrm{ref}}(\lambda)
\big\}_{j=1}^{M_{\mathrm{ref}}}
\]
a refinement basis constructed from predictor-derived geometric cues. This refinement step is necessary because regions requiring additional correction need not coincide exactly with the most active predictor basis functions.

Finally, to preserve robustness and global coverage, we introduce a fixed background basis
\[
\Theta^{\mathrm{bg}}
=
\big\{
\theta_k^{\mathrm{bg}}
\big\}_{k=1}^{M_{\mathrm{bg}}}.
\]

The full corrector dictionary is therefore the enriched basis
\begin{equation}
	\Theta_\lambda^{\mathrm{corr}}
	=
	\Theta_\lambda^{\mathrm{inh}}
	\cup
	\Theta_\lambda^{\mathrm{ref}}
	\cup
	\Theta^{\mathrm{bg}}.
	\label{eq:enriched_basis}
\end{equation}
Thus, the corrector receives from the predictor not merely a coarse field value, but a task-adaptive hidden-basis structure consisting of inherited active kernels and predictor-informed refinement cues.

The corrected field is represented as
\begin{equation}
	u^{\mathrm{corr}}(\mathbf{z};\lambda)
	=
	\sum_{m=1}^{M_\lambda}
	c_m(\lambda)\,
	\phi\!\big(\mathbf{z};\theta_m^{(\lambda)}\big),
	\qquad
	\theta_m^{(\lambda)}\in\Theta_\lambda^{\mathrm{corr}},
	\label{eq:corrector_ansatz}
\end{equation}
where
\[
M_\lambda = M_{\mathrm{inh}} + M_{\mathrm{ref}} + M_{\mathrm{bg}},
\]
and \(c_m(\lambda)\) are task-specific output coefficients.

Since the governing PDE family is linear, enforcing the residual together with the relevant boundary and/or initial conditions at collocation points yields a linear least-squares system
\begin{equation}
	A_\lambda\,c(\lambda)\approx b_\lambda,
	\label{eq:linear_ls_system}
\end{equation}
where the design matrix \(A_\lambda\) is assembled by applying the differential operators to the frozen basis functions in \eqref{eq:corrector_ansatz}. The corrector coefficients are then obtained as
\begin{equation}
	c(\lambda)
	=
	\arg\min_{c\in\mathbb{R}^{M_\lambda}}
	\|A_\lambda c-b_\lambda\|_2^2.
	\label{eq:corrector_ls}
\end{equation}

\paragraph{Role of the two components.}
The KAPI predictor learns, across tasks, where expressive basis support should be placed. The corrector then solves, for each task, how those basis functions should be linearly combined to satisfy the governing physics in the enriched predictor-guided basis.

\paragraph{Overall pipeline.}
The resulting solver has the form
\begin{equation}
	\lambda
	\;\longrightarrow\;
	\Gamma_\psi(\lambda)
	\;\longrightarrow\;
	u_\theta^{\mathrm{pred}}(\cdot;\lambda)
	\;\longrightarrow\;
	\Theta_\lambda^{\mathrm{corr}}
	\;\longrightarrow\;
	u^{\mathrm{corr}}(\cdot;\lambda).
	\label{eq:method_pipeline}
\end{equation}
This formulation casts parametric PDE solving as a task-conditioned basis-generation problem, followed by a physics-constrained coefficient recovery in an enriched predictor-guided basis.


\section{Test Cases}
\label{sec:problem_definition}

We consider four families of parametric linear PDEs spanning localized elliptic response, constant-speed transport, mixed transport--diffusion, and variable-speed transport. In each case, the task is indexed by a low-dimensional parameter vector $\lambda$, and the full KAPI predictor uses a problem-specific shallow hypothesis whose basis geometry is generated from $\lambda$ by its internal meta-network.

\subsection{Test Case 1: 2D Poisson with Gaussian source}
\label{subsec:problem_poisson}

Let $\Omega=[0,1]^2$. For task parameter
\[
\lambda = (x_0,y_0,\nu),
\]
we consider a 2D Poisson test case \cite{DWIVEDI2026133090}
\begin{equation}
	-\Delta u(x,y;\lambda)=f(x,y;\lambda),
	\qquad
	(x,y)\in\Omega,
	\label{eq:poisson_pde}
\end{equation}
with homogeneous Dirichlet boundary condition
\begin{equation}
	u(x,y;\lambda)=0,
	\qquad
	(x,y)\in\partial\Omega,
	\label{eq:poisson_bc}
\end{equation}
and Gaussian source
\begin{equation}
	f(x,y;\lambda)
	=
	\frac{1}{2\pi \nu^2}
	\exp\!\left(
	-\frac{(x-x_0)^2+(y-y_0)^2}{2\nu^2}
	\right).
	\label{eq:poisson_source}
\end{equation}

In the current implementation, the training parameters are sampled from
\begin{equation}
	x_0 \in [0.4,0.6],
	\qquad
	y_0 \in [0.4,0.6],
	\qquad
	\nu \in [0.05,0.10],
	\label{eq:poisson_ranges}
\end{equation}
with $\nu$ sampled log-uniformly.

The predictor uses the hypothesis
\begin{equation}
	u^{\mathrm{pred}}(x,y;\lambda)
	=
	x(1-x)y(1-y)
	\sum_{j=1}^{M}
	g_j(\lambda)\,c_j\,
	\exp\!\left(
	-\frac{(x-\mu_{x,j}(\lambda))^2+(y-\mu_{y,j}(\lambda))^2}
	{\sigma_j(\lambda)^2}
	\right),
	\label{eq:poisson_predictor_hypothesis}
\end{equation}
where the internal meta-network predicts the gates $g_j(\lambda)$, centers $\mu_{x,j}(\lambda),\mu_{y,j}(\lambda)$, and widths $\sigma_j(\lambda)$, while the coefficients $\{c_j\}_{j=1}^M$ are shared across tasks. The prefactor $x(1-x)y(1-y)$ hard-enforces the boundary condition.

\subsection{Test Case 2: Periodic constant-coefficient linear advection}
\label{subsec:problem_advection}

Let $\Omega=[0,1)\times[0,1]$. For task parameter
\[
\lambda=(x_0,\nu),
\]
we consider the standard Gaussian profile transport case \cite{leveque2007finite}  
\begin{equation}
	u_t + u_x = 0,
	\qquad
	(x,t)\in [0,1)\times[0,1],
	\label{eq:adv_pde}
\end{equation}
with periodic boundary condition
\begin{equation}
	u(0,t;\lambda)=u(1,t;\lambda),
	\label{eq:adv_bc}
\end{equation}
and periodic Gaussian initial condition
\begin{equation}
	u(x,0;\lambda)
	=
	\exp\!\left(
	-\frac{d_{\mathrm{per}}(x,x_0)^2}{2\nu^2}
	\right),
	\label{eq:adv_ic}
\end{equation}
where $d_{\mathrm{per}}$ denotes the wrapped distance on the periodic interval.

The training parameters are sampled from
\begin{equation}
	x_0 \in [0.2,0.8],
	\qquad
	\nu \in [0.03,0.12],
	\label{eq:adv_ranges}
\end{equation}
with $\nu$ sampled log-uniformly.

The predictor uses a dynamic periodic Gaussian basis:
\begin{equation}
	u^{\mathrm{pred}}(x,t;\lambda)
	=
	\sum_{j=1}^{M}
	g_j(\lambda)\,\alpha_j(t,\lambda)\,
	\exp\!\left(
	-\frac{d_{\mathrm{per}}(x,\xi_j(t,\lambda))^2}{2h_j(t,\lambda)^2}
	\right),
	\label{eq:adv_predictor_hypothesis}
\end{equation}
where the internal meta-network generates the task-level gates $g_j(\lambda)$, while the full KAPI predictor produces the dynamic amplitudes $\alpha_j(t,\lambda)$, moving centers $\xi_j(t,\lambda)$, and widths $h_j(t,\lambda)$ through a time-dependent task-conditioned basis evolution.

\subsection{Test Case 3: Advection--diffusion}
\label{subsec:problem_advecdiff}

Let $\Omega=[0,1]\times[0,0.5]$. For task parameter
\[
\lambda=(a,\nu),
\]
we solve the 1D version of the standard mixed advection-diffusion test case \cite{BORKER2017520}
\begin{equation}
	u_t + a\,u_x = \nu u_{xx},
	\qquad
	(x,t)\in [0,1]\times[0,0.5].
	\label{eq:advecdiff_pde}
\end{equation}
with initial condition
\begin{equation}
	u(x,0;\lambda)=\exp\!\left(-\frac{(x-x_c)^2}{\nu}\right),
	\qquad
	x_c=0.2.
	\label{eq:advecdiff_ic}
\end{equation}

In the current implementation, the reference family used to generate benchmark traces is
\begin{equation}
	u(x,t;a,\nu)
	=
	\frac{1}{\sqrt{4t+1}}
	\exp\!\left(
	-\frac{(x-x_c-at)^2}{\nu(4t+1)}
	\right).
	\label{eq:advecdiff_exact_family}
\end{equation}

The training parameters are sampled from
\begin{equation}
	a \in [0.5,1.0],
	\qquad
	\nu \in [0.01,0.05],
	\label{eq:advecdiff_ranges}
\end{equation}
with $\nu$ sampled log-uniformly. A viscosity curriculum is used during training: in the first half of training, $\nu$ is sampled from $[0.03,0.05]$, and in the second half from the full range $[0.01,0.05]$.

The predictor uses the residual-corrective ansatz
\begin{equation}
	u^{\mathrm{pred}}(x,t;\lambda)
	=
	u_0(x;\nu)
	+
	t\sum_{j=1}^{M}
	\alpha_j(t,\lambda)\,
	\exp\!\left(
	-\frac{(x-\xi_j(t,\lambda))^2}{2h_j(t,\lambda)^2}
	\right),
	\label{eq:advecdiff_predictor_hypothesis}
\end{equation}
where
\[
u_0(x;\nu)=\exp\!\left(-\frac{(x-x_c)^2}{\nu}\right).
\]
The dynamic quantities $\alpha_j(t,\lambda)$, $\xi_j(t,\lambda)$, and $h_j(t,\lambda)$ are generated within the full KAPI predictor by a time- and parameter-conditioned network, while the residual-corrective form builds the initial condition directly into the predictor ansatz.
\subsection{Test Case 4: Variable-speed advection}
\label{subsec:problem_varadv}

Let $\Omega=[0,1)\times[0,1]$. For task parameter
\[
\lambda=(x_0,\nu,\beta),
\]
we solve
\begin{equation}
	u_t + a(x;\beta)\,u_x = 0,
	\qquad
	a(x;\beta)=1+\beta\sin(2\pi x),
	\label{eq:varadv_pde}
\end{equation}
with periodic boundary condition
\begin{equation}
	u(0,t;\lambda)=u(1,t;\lambda),
	\label{eq:varadv_bc}
\end{equation}
and periodic Gaussian initial condition
\begin{equation}
	u(x,0;\lambda)
	=
	\exp\!\left(
	-\frac{d_{\mathrm{per}}(x,x_0)^2}{2\nu^2}
	\right).
	\label{eq:varadv_ic}
\end{equation}

The training parameters are sampled from
\begin{equation}
	x_0 \in [0.2,0.8],
	\qquad
	\nu \in [0.03,0.12],
	\qquad
	\beta \in [0.20,0.60],
	\label{eq:varadv_ranges}
\end{equation}
with $\nu$ sampled log-uniformly. A width curriculum is used, beginning with broader packets and later expanding to the full range.

The predictor uses
\begin{equation}
	u^{\mathrm{pred}}(x,t;\lambda)
	=
	u_0(x;\lambda)
	+
	t\sum_{j=1}^{M}
	\alpha_j(t,\lambda)\,
	\exp\!\left(
	-\frac{d_{\mathrm{per}}(x,\xi_j(t,\lambda))^2}{2h_j(t,\lambda)^2}
	\right),
	\label{eq:varadv_predictor_hypothesis}
\end{equation}
where
\[
u_0(x;\lambda)
=
\exp\!\left(
-\frac{d_{\mathrm{per}}(x,x_0)^2}{2\nu^2}
\right).
\]
In the current implementation, the dynamic basis inside the full KAPI predictor is generated relative to a learned periodic base dictionary through predicted amplitudes, center shifts, and width perturbations.

\paragraph{Remarks}
\begin{itemize}
	\item \textbf{Inductive bias.}
	Across all four PDE families, the predictor is restricted to a shallow, task-conditioned Gaussian basis whose geometry is generated from the PDE parameter vector $\lambda$. The common inductive bias is that the dominant variation across tasks is largely geometric: localized forcing moves in space, transported packets shift in space-time, widths sharpen or diffuse, and variable-speed transport bends the active solution support. The meta-learner is therefore designed to predict where basis functions should be placed, how wide they should be, and which ones should remain active, while the final physics-consistent coefficient recovery is delegated to the corrector.
	\item  \textbf{Gaussian forcing and initial conditions.}
	Gaussian forcing and Gaussian initial profiles are used throughout because they provide a simple and controllable mechanism for varying two key attributes of the solution family: \emph{locality} and \emph{gradient strength}. Their centers determine where the dominant solution support is expected to emerge, while their widths control how sharply localized the resulting field or transported packet becomes. This makes them particularly suitable for evaluating whether the predictor correctly learns task-dependent basis placement and scale adaptation across elliptic, transport, and mixed regimes.
\end{itemize}
\section{Results and Discussion}
\label{sec:results}

We organize the numerical evaluation around the following questions:
\begin{enumerate}
	\item How well does the meta-learned predictor perform on its own, both quantitatively and through interpretable basis geometry, when compared with parametric neural baselines such as FiLM-HyperPINN and physics-informed DeepONet?
	\item How does the full predictor-corrector framework perform across the four PDE families, both within and beyond the training range, and how strongly does it degrade under extrapolation?
	\item How much of the final gain comes specifically from predictor-guided basis adaptation, as opposed to using a uniform-grid PIELM corrector?
	\item How does the meta-learned solver compare with a standard single-instance PINN in terms of accuracy and computational cost?
\end{enumerate}

Reference solutions are obtained either analytically or through high-resolution numerical solvers, as described in Appendix~\ref{app:ground_truth}, and the principal predictor and corrector implementation details are summarized in Appendix~\ref{app:implementation}. For each test case, we report the relative discrete $L^2$ error on a uniform evaluation grid for both the predictor and the corrected solution:
\begin{equation}
	\mathcal{E}_{\mathrm{pred}}
	=
	\frac{\|u^{\mathrm{pred}}-u_{\mathrm{ref}}\|_2}{\|u_{\mathrm{ref}}\|_2},
	\qquad
	\mathcal{E}_{\mathrm{corr}}
	=
	\frac{\|u^{\mathrm{corr}}-u_{\mathrm{ref}}\|_2}{\|u_{\mathrm{ref}}\|_2}.
	\label{eq:result_error_metrics}
\end{equation}

\subsection{Predictor Performance Against Parametric Neural Baselines}
\label{subsec:predictor_vs_baselines}

We first evaluate the meta-learned predictor on two representative parametric PDE families: the 2D Poisson equation with Gaussian source terms, which is diffusion-dominated, and the 1D periodic linear advection equation with Gaussian initial conditions, which is transport-dominated. In both settings, we compare the \emph{KAPI predictor} against two parametric neural baselines, FiLM-HyperPINN~\cite{perez2018film} and physics-informed DeepONet~\cite{Goswami2023}, using four representative test cases spanning both in-range and out-of-range regimes. The purpose of this comparison is to assess how much physics the predictor alone is able to capture before any corrector is applied. The details of FiLM-HyperPINN and physics-informed Deeponet are provided in Appendix~\ref{app:film_hyper} and ~\ref{app:pideeponet} respectively.

\begin{figure*}[t]
	\centering
	\includegraphics[width=\textwidth]{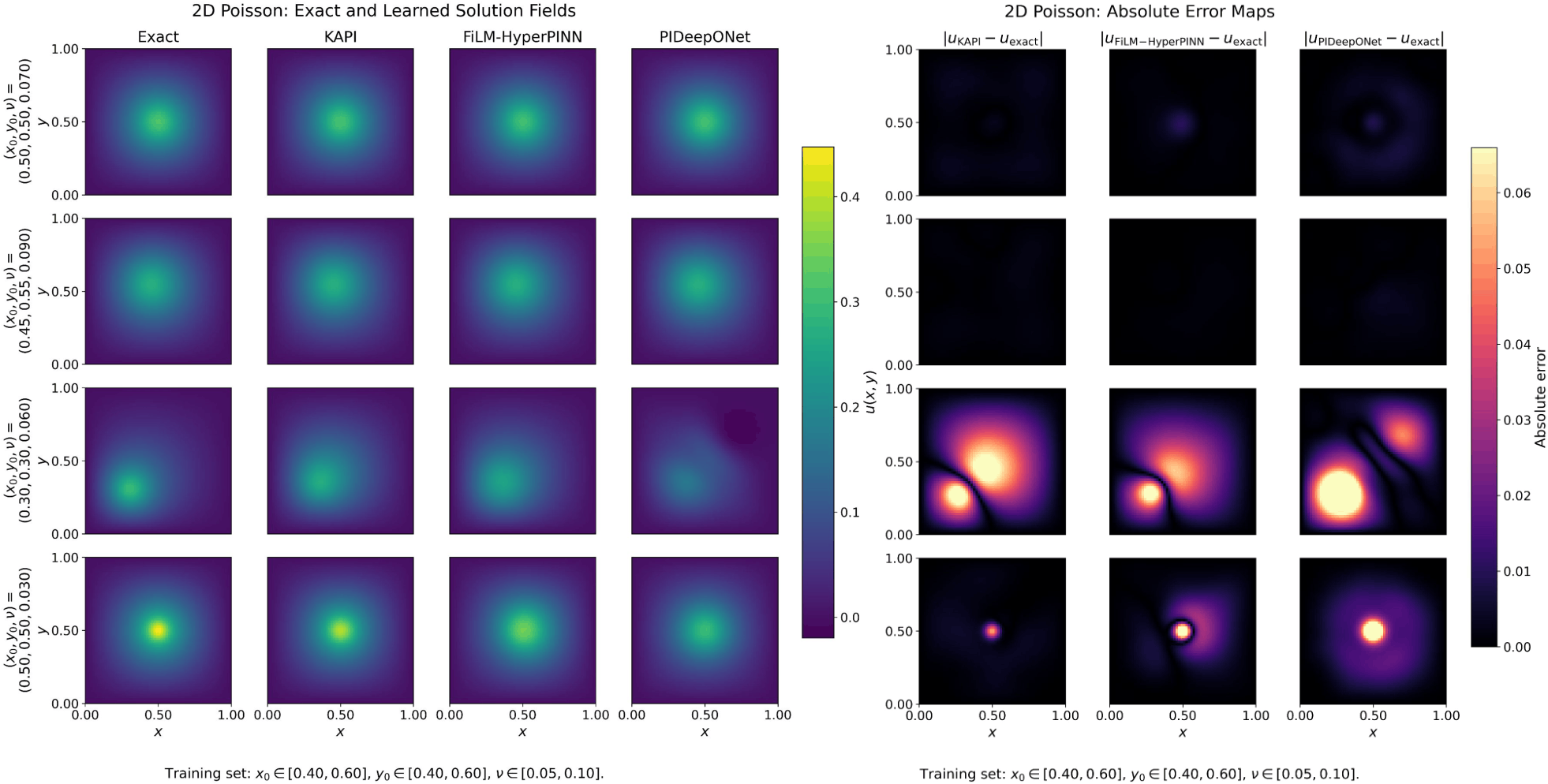}
	\caption{
		Comparison of exact and learned solution fields, together with absolute error maps, for the 2D Poisson problem.
		The left block shows the exact solution and the predictions of the KAPI predictor, FiLM-HyperPINN, and PI-DeepONet.
		The right block shows the corresponding absolute errors.
		The four rows correspond to the test cases $(x_0,y_0,\nu)=(0.50,0.50,0.07)$, $(0.45,0.55,0.09)$, $(0.30,0.30,0.06)$, and $(0.50,0.50,0.03)$.
		The training range is $x_0\in[0.40,0.60]$, $y_0\in[0.40,0.60]$, and $\nu\in[0.05,0.10]$.
	}
	\label{fig:poisson_kapi_hyper_pideeponet}
\end{figure*}

\begin{table}[t]
	\centering
	\caption{Relative $L_2$ errors for the 2D Poisson predictor on four representative test cases.}
	\label{tab:poisson_comparison}
	\setlength{\tabcolsep}{6pt}
	\begin{tabular}{lcccc}
		\toprule
		Test case $(x_0,y_0,\nu)$ & Range type & KAPI predictor & FiLM-HyperPINN & PI-DeepONet \\
		\midrule
		$(0.50,\,0.50,\,0.07)$ & in-range     & $2.008\times10^{-2}$ & $1.657\times10^{-2}$ & $2.946\times10^{-2}$ \\
		$(0.45,\,0.55,\,0.09)$ & in-range     & $1.195\times10^{-2}$ & $6.055\times10^{-3}$ & $1.345\times10^{-2}$ \\
		$(0.30,\,0.30,\,0.06)$ & out-of-range & $3.569\times10^{-1}$ & $2.717\times10^{-1}$ & $4.129\times10^{-1}$ \\
		$(0.50,\,0.50,\,0.03)$ & out-of-range & $3.869\times10^{-2}$ & $9.953\times10^{-2}$ & $1.543\times10^{-1}$ \\
		\bottomrule
	\end{tabular}
\end{table}

\begin{figure*}[t]
	\centering
	\includegraphics[width=\textwidth]{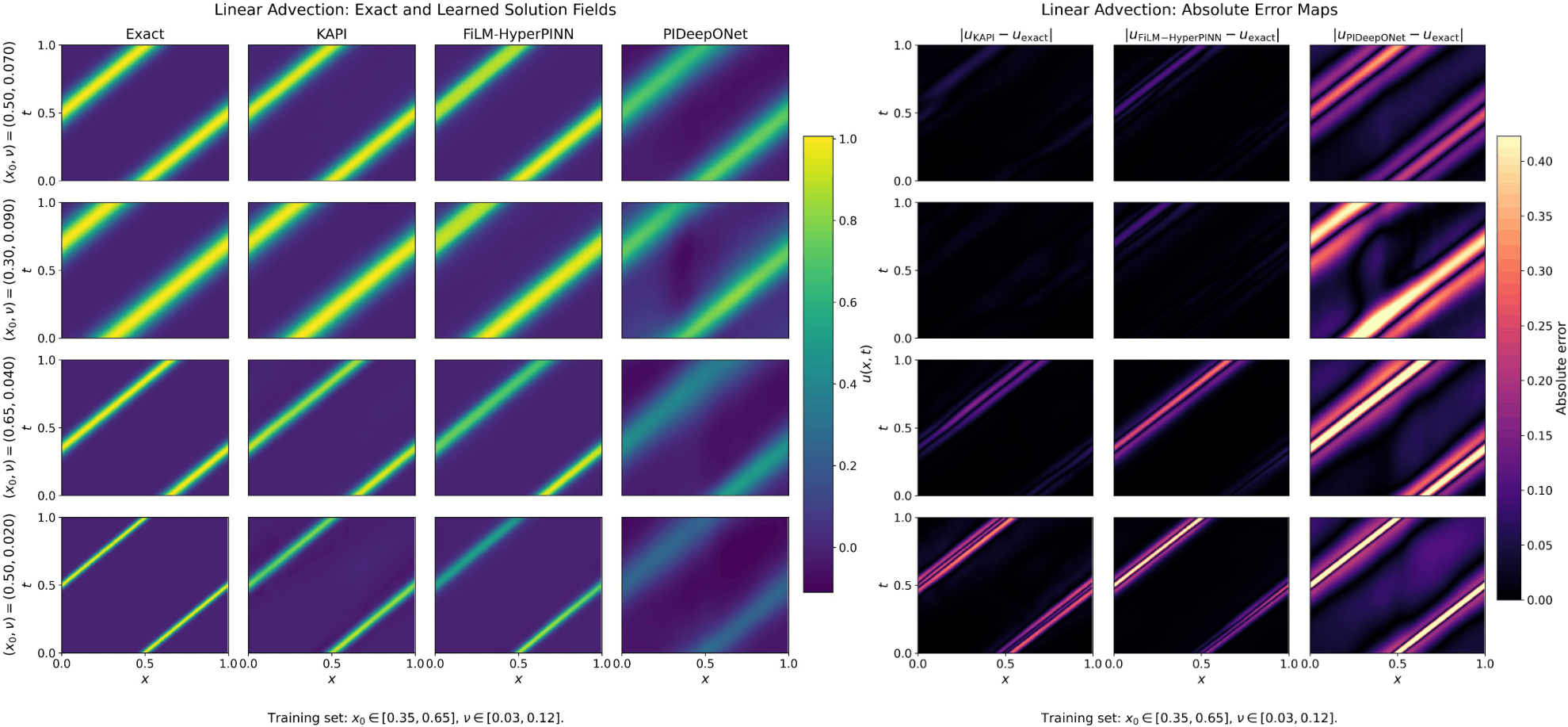}
	\caption{
		Comparison of exact and learned solution fields, together with absolute error maps, for the 1D linear advection problem.
		The left block shows the exact solution and the predictions of the KAPI predictor, FiLM-HyperPINN, and PI-DeepONet.
		The right block shows the corresponding absolute errors.
		The four rows correspond to the test cases $(x_0,\nu)=(0.50,0.07)$, $(0.30,0.09)$, $(0.65,0.04)$, and $(0.50,0.02)$.
		The training range is $x_0\in[0.35,0.65]$ and $\nu\in[0.03,0.12]$.
	}
	\label{fig:advection_kapi_hyper_pideeponet}
\end{figure*}

\begin{table}[t]
	\centering
	\caption{Relative $L_2$ errors for the 1D linear advection predictor on four representative test cases.}
	\label{tab:advection_comparison}
	\setlength{\tabcolsep}{6pt}
	\begin{tabular}{lcccc}
		\toprule
		Test case $(x_0,\nu)$ & Range type & KAPI predictor & FiLM-HyperPINN & PI-DeepONet \\
		\midrule
		$(0.50,\,0.07)$ & in-range      & $4.022\times10^{-2}$ & $5.581\times10^{-2}$ & $2.950\times10^{-1}$ \\
		$(0.65,\,0.04)$ & in-range      & $1.166\times10^{-1}$ & $2.047\times10^{-1}$ & $8.450\times10^{-1}$ \\
		$(0.30,\,0.09)$ & out-of-range  & $2.215\times10^{-2}$ & $4.203\times10^{-2}$ & $1.715\times10^{-1}$ \\
		$(0.50,\,0.02)$ & out-of-range  & $3.656\times10^{-1}$ & $3.752\times10^{-1}$ & $7.383\times10^{-1}$ \\
		\bottomrule
	\end{tabular}
\end{table}

For the Poisson family, all three predictors perform well on the in-range cases, with FiLM-HyperPINN achieving the smallest errors and the KAPI predictor remaining close behind. The differences become more pronounced under extrapolation. In the shifted-source out-of-range case, FiLM-HyperPINN remains slightly stronger, but for the narrow-source case the KAPI predictor is substantially more robust, while PI-DeepONet exhibits the largest degradation overall. Thus, in the diffusion-dominated regime, the KAPI predictor is already competitive with a strong parametric PINN baseline and is notably more reliable than PI-DeepONet when localization becomes sharper than what was seen during training.

The advection family reveals a clearer separation. Here the KAPI predictor achieves the lowest error on all four test cases, including both in-range and out-of-range settings, while FiLM-HyperPINN remains competitive and PI-DeepONet is consistently less accurate. The visual comparisons in Fig.~\ref{fig:advection_kapi_hyper_pideeponet} show that the KAPI predictor and FiLM-HyperPINN both preserve the transported Gaussian packet reasonably well, whereas PI-DeepONet tends to produce visibly more diffused predictions, especially for narrow or extrapolative cases. This difference is particularly strong in the narrow-pulse regime, where preserving sharp localization is essential.

Taken together, these results show that the predictor alone already captures a significant portion of the governing physics. In the Poisson case it identifies the dominant response region associated with the source, while in the advection case it tracks the transported packet with substantially better fidelity than a global operator-style baseline. This is precisely the behavior required for the predictor to serve as the geometry generator for the subsequent corrector.

\subsection{Interpretability of the Meta-Learned Predictor}
\label{subsec:predictor_interpretability}

\begin{figure*}[t]
	\centering
	\begin{subfigure}[t]{0.47\textwidth}
		\centering
		\includegraphics[width=\textwidth]{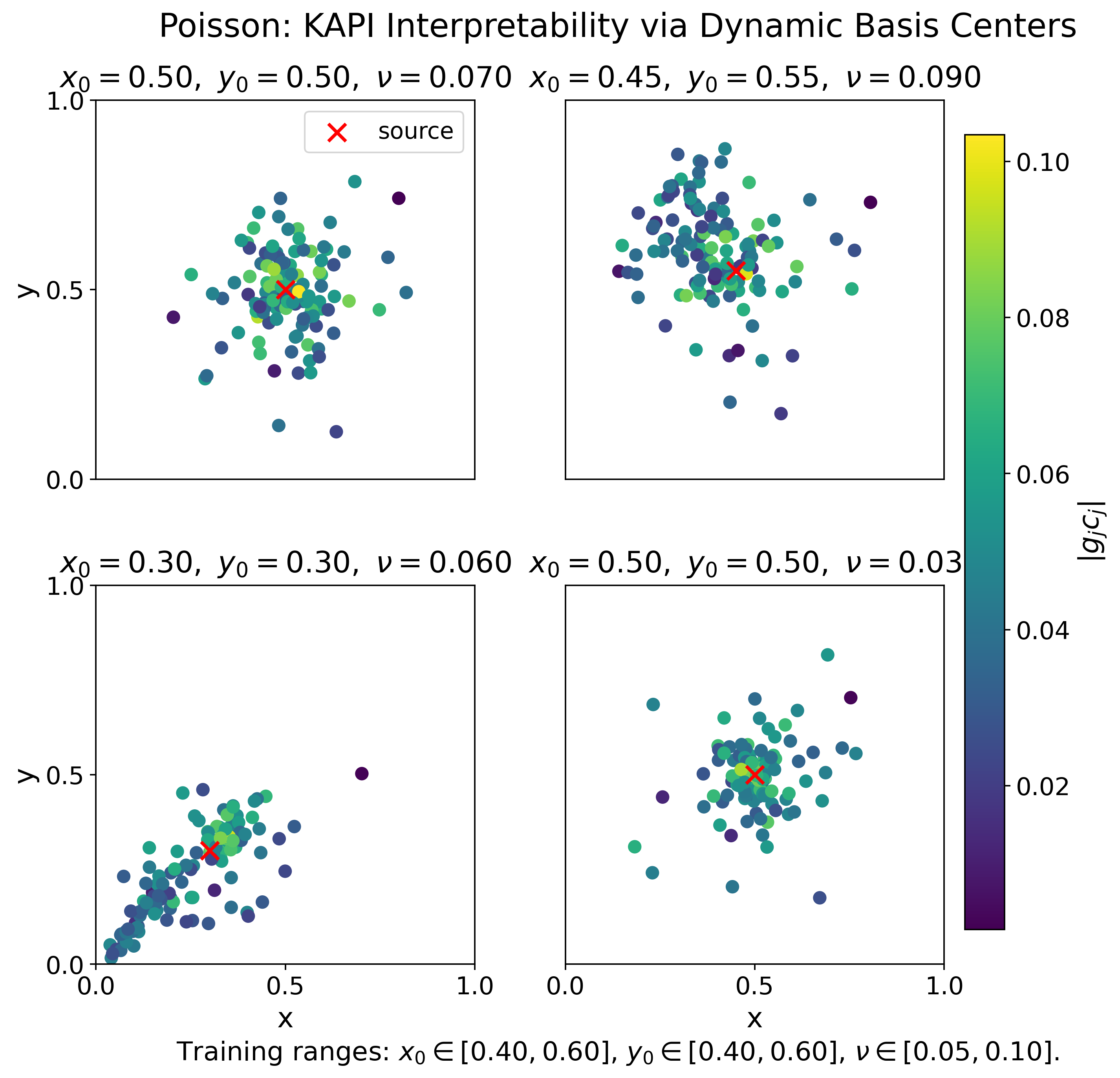}
		\caption{2D Poisson: the learned basis centers cluster around the active source region, with stronger coefficients concentrated near the localized forcing.}
		\label{fig:tc1b}
	\end{subfigure}
	\hfill
	\begin{subfigure}[t]{0.49\textwidth}
		\centering
		\includegraphics[width=\textwidth]{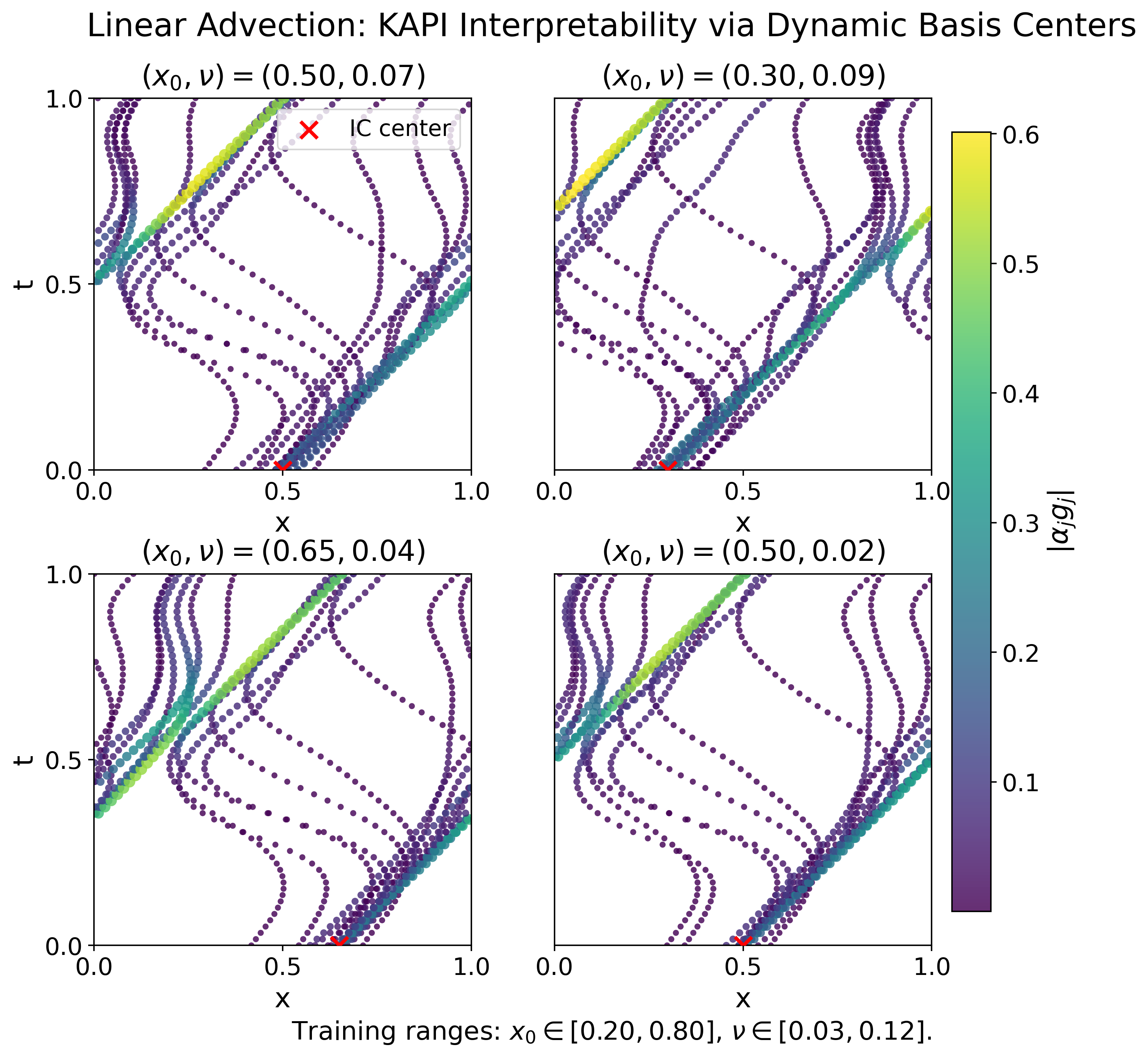}
		\caption{1D linear advection: dynamic basis-center trajectories align with the transported packet, revealing a transport-aware basis geometry.}
		\label{fig:tc2b}
	\end{subfigure}
	\caption{
		Interpretability of the KAPI predictor through task-dependent basis geometry. In Poisson, the learned basis adapts to the source location and localization scale. In linear advection, the most active basis centers move along the dominant transport direction.}
	\label{fig:kapi_interpretability}
\end{figure*}

To understand why the KAPI predictor performs well, we examine the task-dependent RBF geometry generated inside the predictor by its internal meta-network. Figure~\ref{fig:kapi_interpretability} shows that the learned basis configuration is strongly aligned with the dominant solution structure in both PDE families: for Poisson, the active basis centers concentrate around the source region, whereas for linear advection, the most active centers trace characteristic-like trajectories in space-time.

For the Poisson problem, the interpretability plots show that the KAPI predictor adapts its spatial basis toward the source-centered region that dominates the response. When the source remains close to the training regime, the learned centers form a compact cloud around the forcing location. In the narrow-source extrapolation case, the KAPI predictor contracts this cloud appropriately, which explains why it remains robust when extrapolating in $\nu$. By contrast, when the source is shifted farther outside the training range, the learned basis must extrapolate spatially as well, and the resulting alignment is less favorable. This provides a direct geometric explanation for the corresponding increase in error.

For linear advection, the learned geometry becomes explicitly dynamic. The active centers trace trajectories in the $(x,t)$ plane that align closely with the transported packet, showing that the KAPI predictor has learned a transport-aware representation rather than an arbitrary time-dependent parameterization. Even in extrapolative cases, the dominant trajectories remain qualitatively aligned with the correct transport direction. When the pulse becomes narrower than in training, however, the learned widths and center dynamics are not always sharp enough to resolve the packet cleanly, which leads to the larger but still interpretable degradation seen in the narrow out-of-range case.

These plots therefore do more than provide qualitative visualization. They establish that the KAPI predictor is learning the \emph{important regions of the PDE solution manifold}: localized gradient hot spots for Poisson and characteristic-like transport paths for advection. Even when the predictor does not fully resolve those structures, it identifies them reliably enough to provide the task-adaptive basis structure needed by the subsequent corrector. This observation is central to the predictor-corrector philosophy of the paper: the predictor need not be perfect, but it should identify where expressive basis support is required.

\subsection{Predictor-Corrector Performance Across PDE Families}
\label{subsec:pc_performance}
\begin{figure*}[t]
	\centering
	\includegraphics[width=\textwidth]{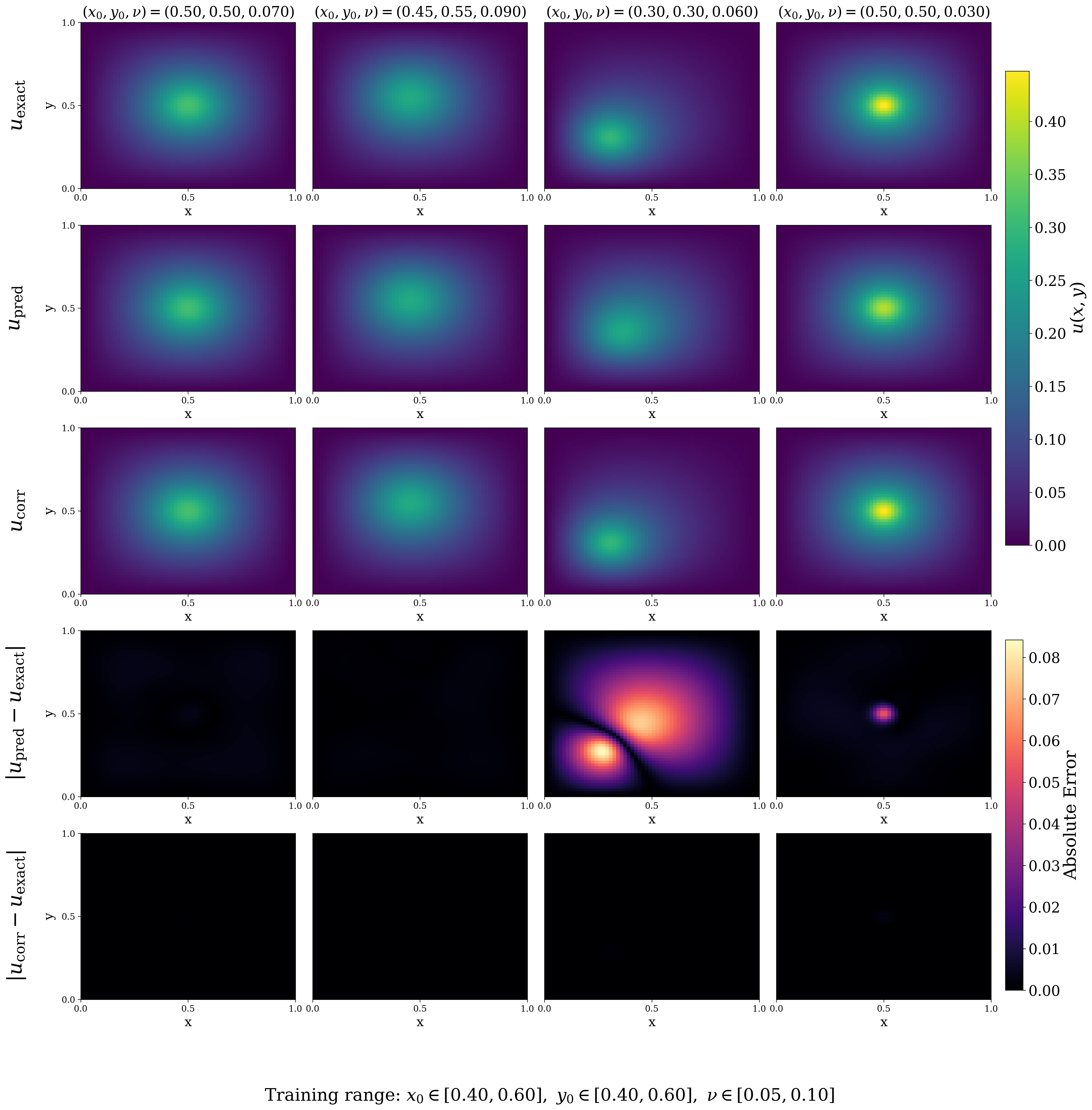}
	\caption{
		Predictor-corrector performance on the 2D Poisson family. Rows show the exact solution $u_{\mathrm{exact}}$, the predictor output $u_{\mathrm{pred}}$, the corrected solution $u_{\mathrm{corr}}$, and the corresponding absolute error maps $|u_{\mathrm{pred}}-u_{\mathrm{exact}}|$ and $|u_{\mathrm{corr}}-u_{\mathrm{exact}}|$. Columns correspond to the test cases $(x_0,y_0,\nu)=(0.50,0.50,0.07)$, $(0.45,0.55,0.09)$, $(0.30,0.30,0.06)$, and $(0.50,0.50,0.03)$. The first two cases lie within the training range, while the latter two probe extrapolation in source location and localization scale.
	}
	\label{fig:tc01_solution}
\end{figure*}

\begin{figure*}[t]
	\centering
	\includegraphics[width=\textwidth]{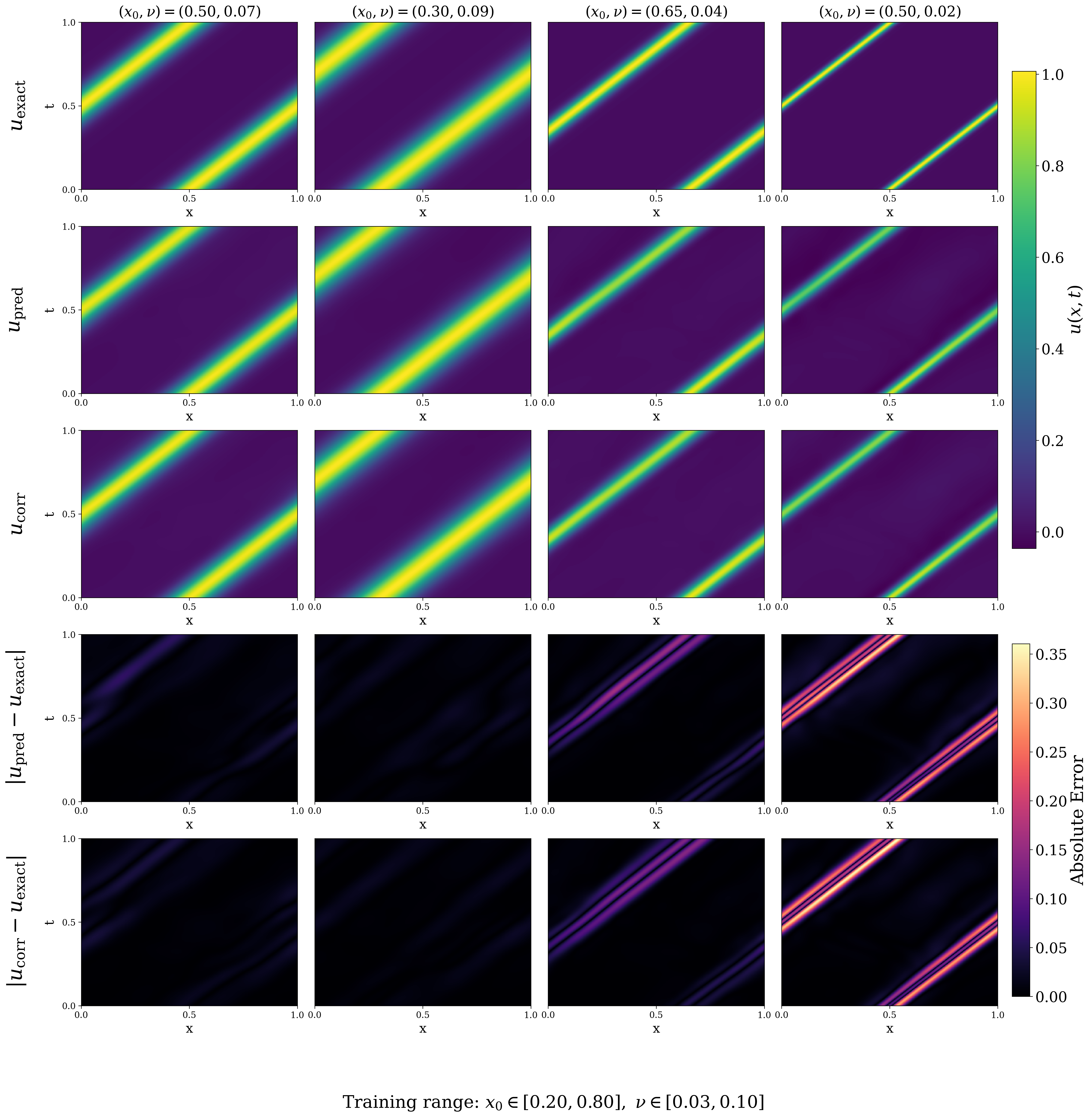}
	\caption{
		Predictor-corrector performance on the 1D periodic linear advection family. Rows show the exact solution $u_{\mathrm{exact}}$, the predictor output $u_{\mathrm{pred}}$, the corrected solution $u_{\mathrm{corr}}$, and the corresponding absolute error maps $|u_{\mathrm{pred}}-u_{\mathrm{exact}}|$ and $|u_{\mathrm{corr}}-u_{\mathrm{exact}}|$. Columns correspond to the test cases $(x_0,\nu)=(0.50,0.07)$, $(0.30,0.09)$, $(0.65,0.04)$, and $(0.50,0.02)$. The first three cases lie within the training range, while the last case probes extrapolation to a narrower packet than seen during training.
	}
	\label{fig:tc02_solution}
\end{figure*}

\begin{figure*}[t]
	\centering
	\includegraphics[width=\textwidth]{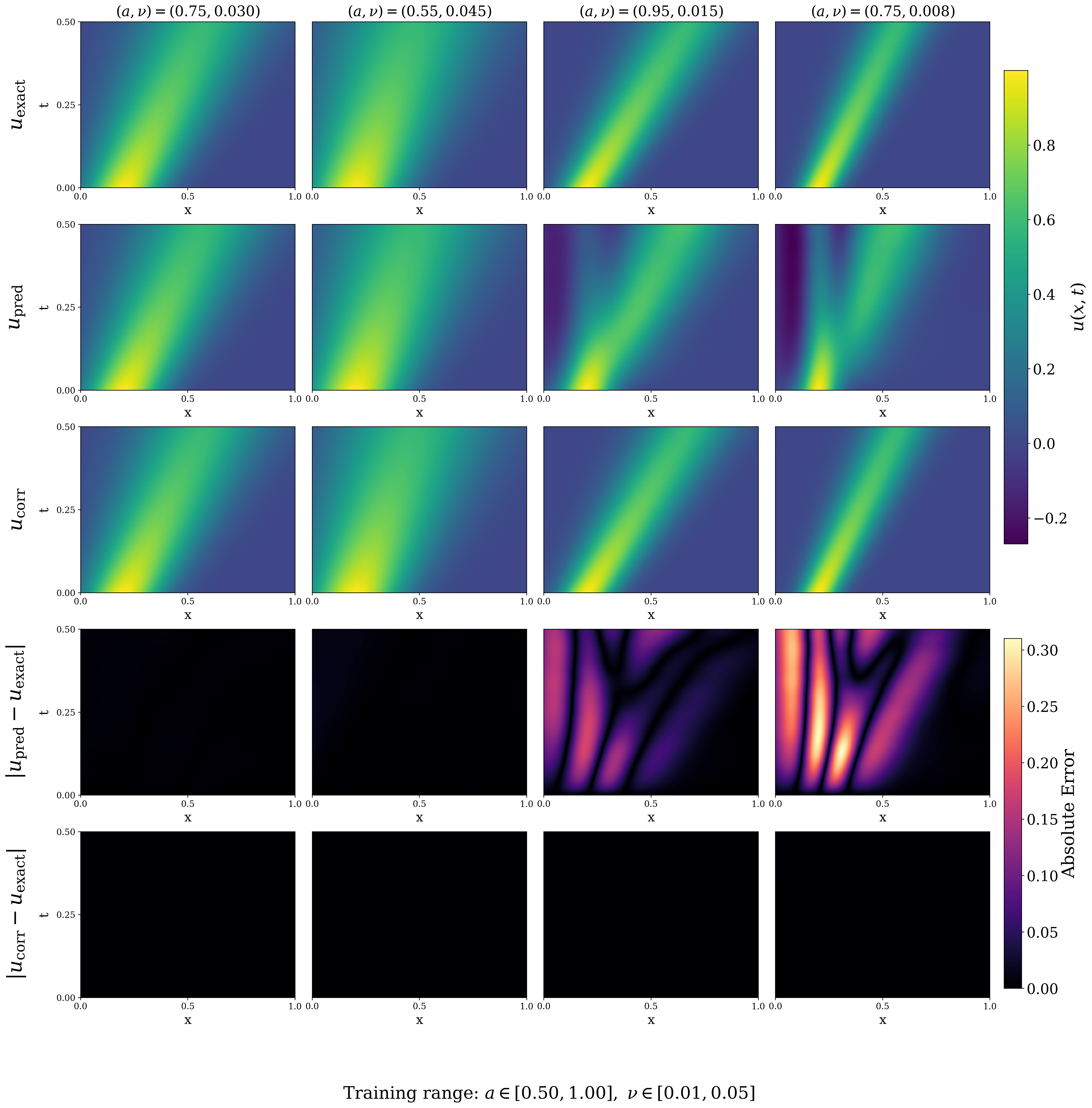}
	\caption{
		Predictor-corrector performance on the advection--diffusion family. Rows show the exact solution $u_{\mathrm{exact}}$, the predictor output $u_{\mathrm{pred}}$, the corrected solution $u_{\mathrm{corr}}$, and the corresponding absolute error maps $|u_{\mathrm{pred}}-u_{\mathrm{exact}}|$ and $|u_{\mathrm{corr}}-u_{\mathrm{exact}}|$. Columns correspond to the test cases $(a,\nu)=(0.75,0.03)$, $(0.55,0.045)$, $(0.95,0.015)$, and $(0.75,0.008)$. The first three cases lie within the training range, while the last case probes extrapolation to a lower-viscosity regime than seen during training.
	}
	\label{fig:tc03_solution}
\end{figure*}

\begin{figure*}[t]
	\centering
	\includegraphics[width=\textwidth]{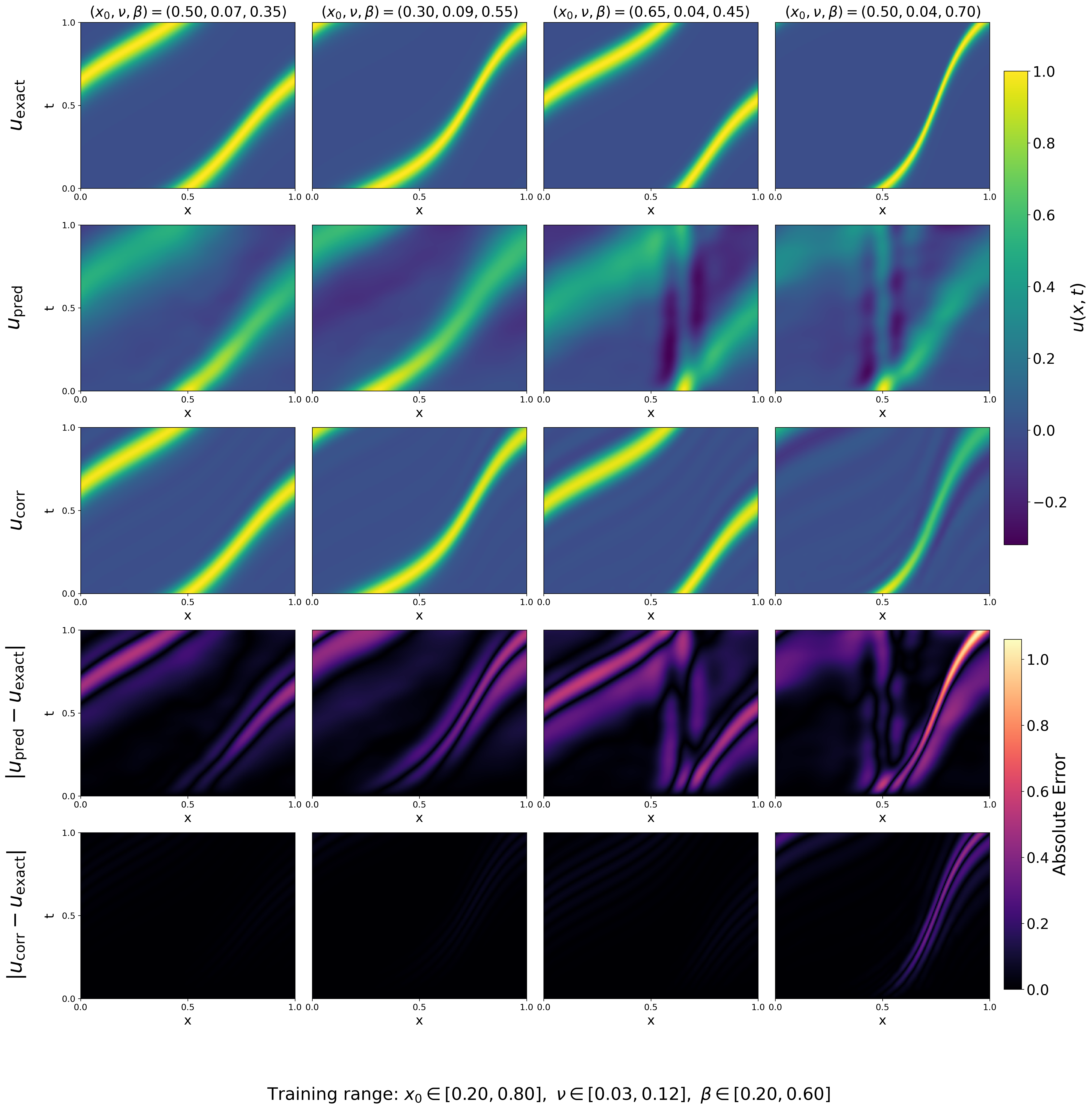}
	\caption{
		Predictor-corrector performance on the variable-speed advection family. Rows show the exact solution $u_{\mathrm{exact}}$, the predictor output $u_{\mathrm{pred}}$, the corrected solution $u_{\mathrm{corr}}$, and the corresponding absolute error maps $|u_{\mathrm{pred}}-u_{\mathrm{exact}}|$ and $|u_{\mathrm{corr}}-u_{\mathrm{exact}}|$. Columns correspond to the test cases $(x_0,\nu,\beta)=(0.50,0.07,0.35)$, $(0.30,0.09,0.55)$, $(0.65,0.04,0.45)$, and $(0.50,0.04,0.70)$. The first three cases lie within the training range, while the last case probes extrapolation to a stronger velocity modulation than seen during training.
	}
	\label{fig:tc04_solution}
\end{figure*}

\begin{table*}[t]
	\centering
	\caption{Predictor and corrector errors across the four PDE families. The corrector typically reduces the error by one or more orders of magnitude, including several extrapolative cases. The main exception is constant-coefficient linear advection, where the predictor already captures the dominant transport geometry and the correction is correspondingly modest.}
	\label{tab:pc_all_cases}
	\small
	\setlength{\tabcolsep}{4pt}
	\renewcommand{\arraystretch}{1.08}
	\begin{tabular}{p{1.55cm} l c c c}
		\toprule
		PDE family & Test case & Regime & $\mathcal{E}_{\mathrm{pred}}$ & $\mathcal{E}_{\mathrm{corr}}$ \\
		\midrule
		\multirow{4}{*}{\shortstack{2D\\Poisson}}
		& $(0.50,\,0.50,\,0.070)$ & in-range        & $2.008\times10^{-2}$ & $7.032\times10^{-4}$ \\
		& $(0.45,\,0.55,\,0.090)$ & in-range        & $1.195\times10^{-2}$ & $5.596\times10^{-4}$ \\
		& $(0.30,\,0.30,\,0.060)$ & out-of-range    & $3.569\times10^{-1}$ & $1.122\times10^{-3}$ \\
		& $(0.50,\,0.50,\,0.030)$ & out-of-range    & $3.869\times10^{-2}$ & $1.656\times10^{-3}$ \\
		\midrule
		\multirow{4}{*}{\shortstack{Linear\\advection}}
		& $(0.50,\,0.070)$        & in-range        & $4.022\times10^{-2}$ & $3.428\times10^{-2}$ \\
		& $(0.30,\,0.090)$        & in-range        & $2.215\times10^{-2}$ & $2.040\times10^{-2}$ \\
		& $(0.65,\,0.040)$        & narrow in-range & $1.166\times10^{-1}$ & $1.153\times10^{-1}$ \\
		& $(0.50,\,0.020)$        & out-of-range    & $3.656\times10^{-1}$ & $3.885\times10^{-1}$ \\
		\midrule
		\multirow{4}{*}{\shortstack{Advection--\\diffusion}}
		& $(0.75,\,0.030)$        & in-range             & $7.644\times10^{-3}$ & $1.959\times10^{-4}$ \\
		& $(0.55,\,0.045)$        & in-range             & $1.055\times10^{-2}$ & $2.225\times10^{-4}$ \\
		& $(0.95,\,0.015)$        & trans.-dom.\ in-range & $1.924\times10^{-1}$ & $1.846\times10^{-4}$ \\
		& $(0.75,\,0.008)$        & out-of-range         & $3.904\times10^{-1}$ & $9.882\times10^{-4}$ \\
		\midrule
		\multirow{4}{*}{\shortstack{Variable-\\speed\\advection}}
		& $(0.50,\,0.070,\,0.350)$ & in-range        & $4.111\times10^{-1}$ & $2.673\times10^{-2}$ \\
		& $(0.30,\,0.090,\,0.550)$ & in-range        & $5.614\times10^{-1}$ & $4.484\times10^{-2}$ \\
		& $(0.65,\,0.040,\,0.450)$ & narrow in-range & $6.249\times10^{-1}$ & $5.715\times10^{-2}$ \\
		& $(0.50,\,0.040,\,0.700)$ & out-of-range    & $1.075\times10^{0}$ & $3.896\times10^{-1}$ \\
		\bottomrule
	\end{tabular}
\end{table*}

We now evaluate the full predictor-corrector framework on all four PDE families, using representative in-range and out-of-range test cases for each problem. Table~\ref{tab:pc_all_cases} summarizes the errors of the KAPI predictor and the corrected solution, while Figs.~\ref{fig:tc01_solution}--\ref{fig:tc04_solution} show the corresponding solution fields and error maps.

A clear pattern emerges. For the 2D Poisson, advection--diffusion, and variable-speed advection families, the corrector produces substantial gains over the predictor, often reducing the relative $L^2$ error by one or more orders of magnitude. In the Poisson family, this includes both in-range cases and the shifted and narrow extrapolative cases, with the corrected errors consistently reduced to the $10^{-3}$ level or below. Extensions of the Poisson test case to two-source and four-source forcing configurations are provided in Appendix~\ref{app:poisson_extend}. The improvement is even more striking for advection--diffusion, where the predictor can degrade substantially in transport-dominated or low-viscosity regimes, yet the corrector still recovers near-reference-quality solutions with errors around $10^{-4}$ to $10^{-3}$. For variable-speed advection, the predictor alone is quantitatively weak for in range or interpolated parameter values, but the corrector remains highly beneficial, reducing errors from $O(10^{-1})$--$O(1)$ down to $O(10^{-2})$--$O(10^{-1})$ in all but the strongest extrapolative case.

The linear advection family is the main exception. There, the predictor already captures the dominant transport geometry well, and the correction step yields only modest gains for broad packets, negligible gains for narrower in-range packets, and a slight degradation in the hardest narrow out-of-range case. This suggests that, for constant-coefficient advection, the main challenge is not locating the transported structure, but resolving sufficiently sharp packet widths once they fall outside the training regime. In such cases, a corrector built from predictor-guided basis structure remains stable but cannot recover resolution that is absent from the basis support generated by the predictor. To assess whether the advection predictor is tied to Gaussian initial data, we also report an additional test with a non-Gaussian Mexican-hat initial condition in Appendix~\ref{app:mexhat_advection}.

Taken together, these results support the central premise of the method. The predictor need not be fully accurate as a stand-alone solver; instead, it must identify a geometrically informative approximation space and transfer a useful predictor-guided basis structure from which the corrector can recover the final solution. The results show that this works especially well for localized elliptic structure, mixed transport--diffusion, and nonlinear transport geometry, and remains stable even when extrapolation degrades predictor quality. The only regime in which the gain becomes limited is when the predictor geometry captures the correct transport direction but lacks the sharpness required to resolve fine-scale features.

\subsection{Why Predictor-Corrector Works: Geometry of the Enriched Basis}
\label{subsec:pc_geometry}

\begin{figure*}[t]
	\centering
	\begin{subfigure}[t]{0.99\textwidth}
		\centering
		\includegraphics[width=\textwidth]{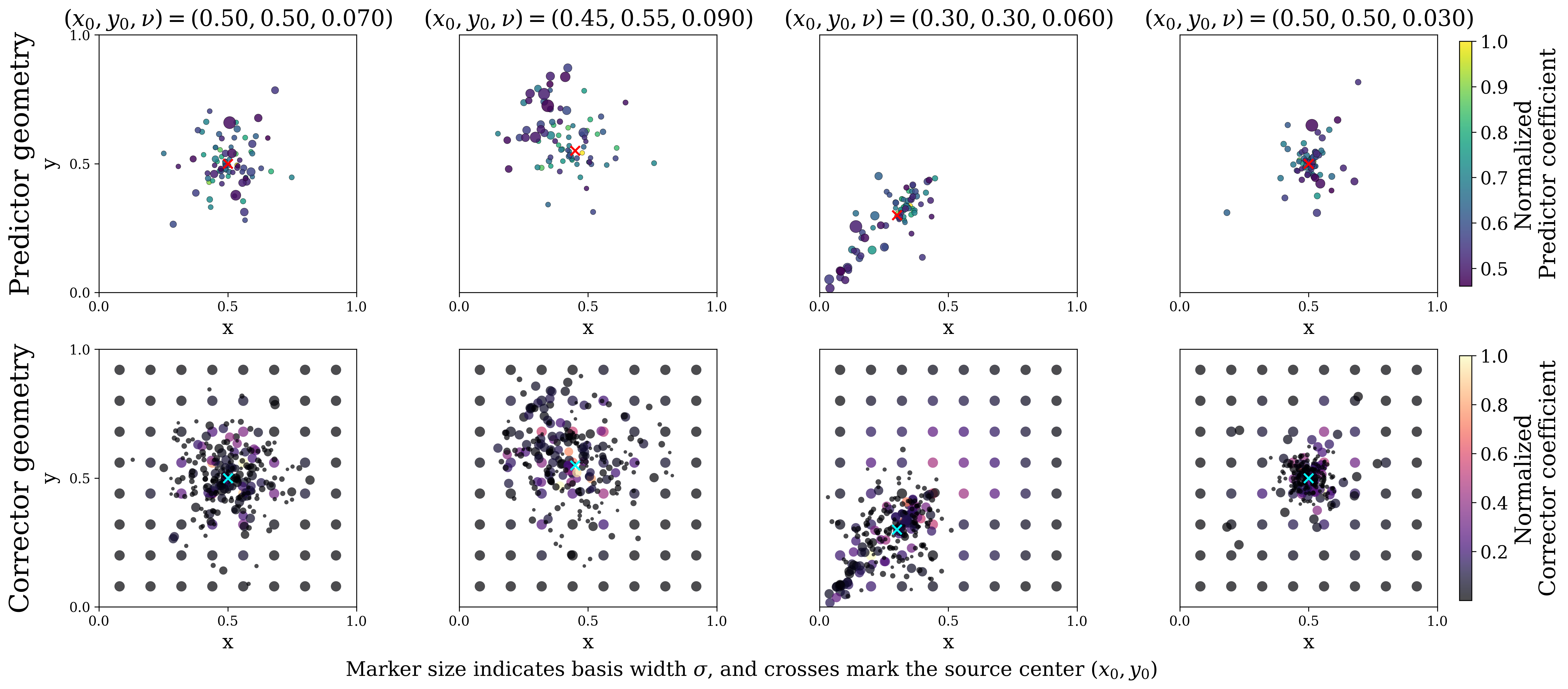}
		\caption{2D Poisson: predictor and corrector basis geometries.}
		\label{fig:pc_geometry_poisson}
	\end{subfigure}
	
	\vspace{0.75em}
	
	\begin{subfigure}[t]{0.99\textwidth}
		\centering
		\includegraphics[width=\textwidth]{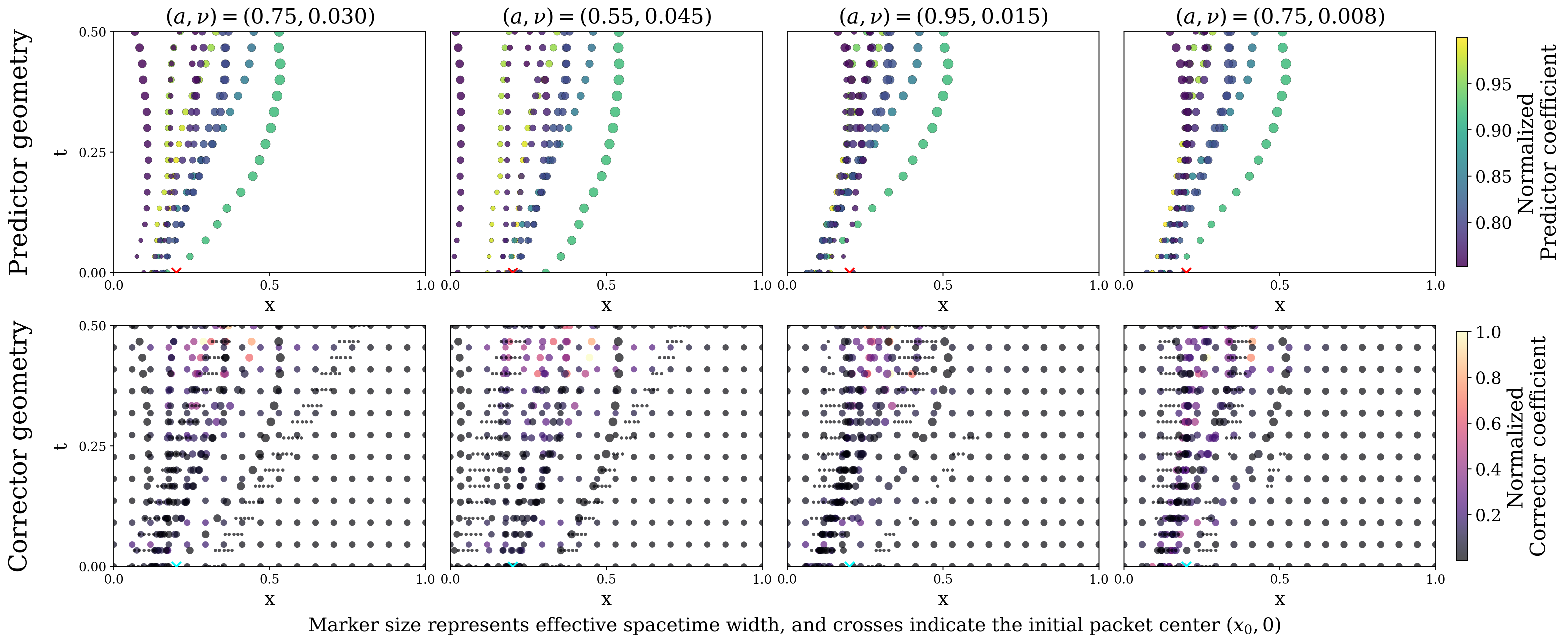}
		\caption{Advection--diffusion: predictor and corrector basis geometries.}
		\label{fig:pc_geometry_advecdiff}
	\end{subfigure}
	\caption{
		Interpretability of the predictor-corrector mechanism through basis geometry. In each panel, the top row shows the predictor geometry, while the bottom row shows the enriched corrector geometry. Marker size indicates basis width, and color indicates normalized coefficient magnitude. The predictor supplies a task-adaptive concentration of basis support near the important region, while the corrector augments this inherited basis with additional refinement and background support, and then recomputes the final coefficients through least squares.
	}
	\label{fig:pc_geometry}
\end{figure*}

To understand why the predictor-corrector construction is effective, we examine the basis geometry before and after correction for two representative cases: 2D Poisson and advection--diffusion. We focus on these two families because they exhibit the clearest predictor-to-corrector gains and therefore best reveal the mechanism of the method. Figure~\ref{fig:pc_geometry} shows the predictor geometry together with the enriched corrector geometry used in the final least-squares solve. In both cases, the predictor supplies a task-adaptive set of basis functions concentrated near the dynamically important region, while the corrector augments this inherited structure with additional refinement and background support to improve local resolution and preserve global coverage.

For the Poisson family, the predictor places its basis functions near the source-centered response region, but the resulting support remains relatively sparse and localized. The corrector then enriches this predictor-guided basis by adding additional support around the source together with a structured background grid over the full domain. This combination explains the strong gains observed in Table~\ref{tab:pc_all_cases}: the predictor already identifies \emph{where} the important elliptic response lives, and the corrector supplements this with enough local and global support to satisfy the PDE accurately everywhere. In the shifted and narrow cases, the same mechanism remains visible: even when the predictor alone is imperfect, its geometry still provides a highly informative local prior, and the enriched corrector geometry converts that prior into a high-accuracy final solution.

The advection--diffusion case reveals the same principle in a space-time setting. The predictor generates a dynamic basis aligned with the transported packet, but in the harder low-viscosity cases this basis alone is too coarse to resolve the full structure sharply. The corrector again augments the predictor-guided basis with additional predictor-informed refinement and a broader space-time scaffold, producing a richer hidden dictionary that still remains concentrated near the transported packet. This explains why the corrector is especially effective in the transport-dominated and out-of-range low-viscosity regimes: the predictor identifies the relevant space-time corridor, and the corrector then resolves that corridor much more accurately by recombining an enriched set of basis functions.

These plots clarify the distinct roles of the two components. The predictor is responsible for identifying a task-dependent region of expressive support, while the corrector is responsible for enriching that support and converting it into an accurate physics-consistent approximation. In this sense, the success of the framework does not require the predictor itself to be numerically precise everywhere; it only requires the predictor-guided basis structure to be sufficiently aligned with the important solution structure so that the enriched least-squares corrector can exploit it.

\subsection{Ablation: Is the Gain Due to Predictor-Guided Basis Adaptation?}
\label{subsec:ablation_uniform_vs_guided}

A natural question is whether the final improvement comes from predictor-guided basis adaptation itself, or simply from attaching a PIELM-style least-squares corrector to the end of the pipeline. To answer this, we compare the proposed predictor-guided corrector against a \emph{uniform-only} PIELM corrector that solves the same least-squares physics problem but uses only a fixed background grid of RBFs, without any predictor-guided hidden units. We perform this ablation on two representative PDE families, 2D Poisson and advection--diffusion, since these are the cases where the predictor-corrector gains are strongest. Importantly, the uniform-only baseline is not evaluated with fixed oversized widths: as the background grid is refined, the associated background RBF widths shrink with the grid spacing. Thus, the baseline is allowed to benefit from both increased basis count and increased locality.

\begin{table*}[t]
	\centering
	\caption{Ablation study comparing the predictor-guided corrector against a uniform-only PIELM corrector. For each case, the uniform-only result reports the best error obtained over the tested background-grid sweep. Across both PDE families, the predictor-guided corrector is consistently more accurate, often by one to three orders of magnitude.}
	\label{tab:ablation_uniform_vs_guided}
	\small
	\setlength{\tabcolsep}{4pt}
	\renewcommand{\arraystretch}{1.08}
	\begin{tabular}{p{1.55cm} l c c c}
		\toprule
		PDE family & Test case & Regime & Guided corrector & Uniform-only corrector \\
		\midrule
		\multirow{4}{*}{\shortstack{2D\\Poisson}}
		& $(0.50,\,0.50,\,0.070)$ & in-range        & $7.032\times10^{-4}$ & $1.138\times10^{-2}$ \\
		& $(0.45,\,0.55,\,0.090)$ & in-range        & $5.596\times10^{-4}$ & $2.659\times10^{-2}$ \\
		& $(0.30,\,0.30,\,0.060)$ & out-of-range    & $1.122\times10^{-3}$ & $4.425\times10^{-2}$ \\
		& $(0.50,\,0.50,\,0.030)$ & out-of-range    & $1.656\times10^{-3}$ & $3.445\times10^{-1}$ \\
		\midrule
		\multirow{4}{*}{\shortstack{Advection--\\diffusion}}
		& $(0.75,\,0.030)$        & in-range              & $1.959\times10^{-4}$ & $1.126\times10^{-2}$ \\
		& $(0.55,\,0.045)$        & in-range              & $2.225\times10^{-4}$ & $3.640\times10^{-3}$ \\
		& $(0.95,\,0.015)$        & trans.-dom.\ in-range & $1.846\times10^{-4}$ & $2.380\times10^{-2}$ \\
		& $(0.75,\,0.008)$        & out-of-range          & $9.882\times10^{-4}$ & $3.704\times10^{-2}$ \\
		\bottomrule
	\end{tabular}
\end{table*}

\begin{figure*}[t]
	\centering
	\includegraphics[width=\textwidth]{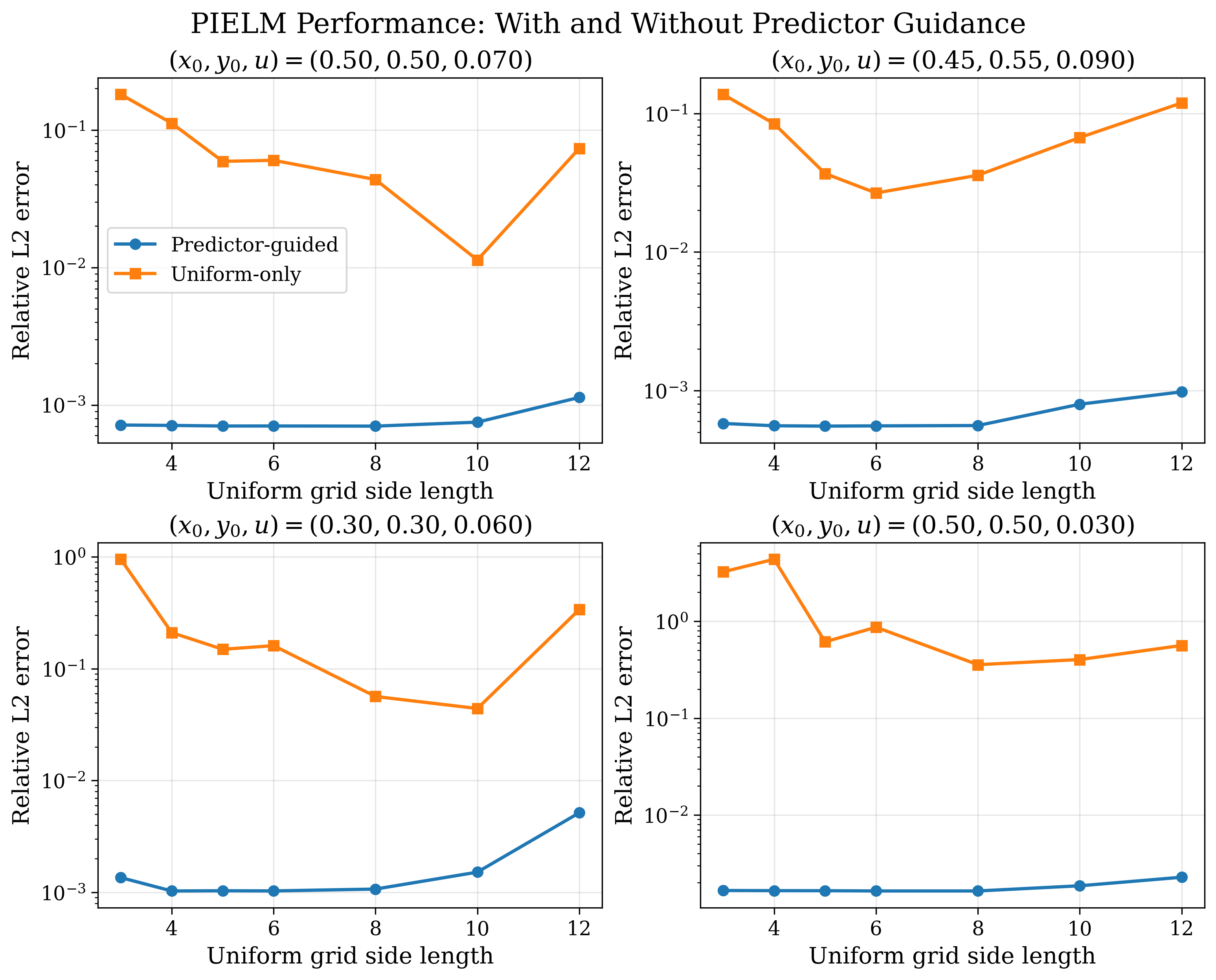}
	\caption{
		Ablation study on the 2D Poisson family. We compare the predictor-guided corrector against a uniform-only PIELM corrector as the background grid side length is varied. Across all four cases, the predictor-guided corrector remains substantially more accurate, and the gap is especially large in the shifted and narrow out-of-range settings. This shows that increasing uniform-grid density alone does not recover the benefit of predictor-guided basis placement.
	}
	\label{fig:ablation_poisson}
\end{figure*}

\begin{figure*}[t]
	\centering
	\includegraphics[width=\textwidth]{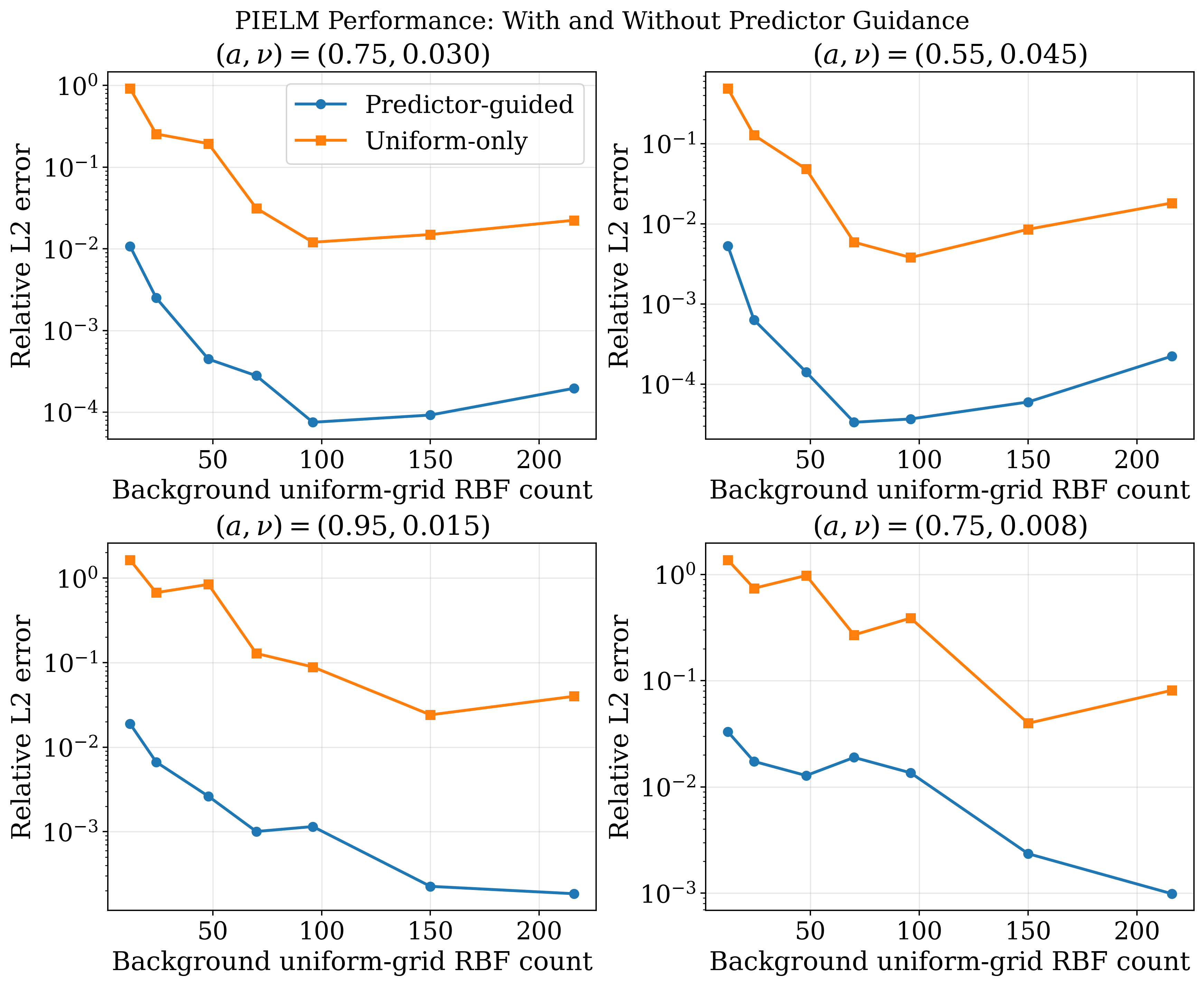}
	\caption{
		Ablation study on the advection--diffusion family. We compare the predictor-guided corrector against a uniform-only PIELM corrector as the background space-time RBF count is varied. The predictor-guided corrector consistently achieves much lower error, particularly in the transport-dominated and low-viscosity regimes. Thus, the main gain comes from task-adaptive basis geometry rather than from least-squares correction alone.
	}
	\label{fig:ablation_advecdiff}
\end{figure*}

The results are unambiguous. In both PDE families, the predictor-guided corrector is consistently and substantially more accurate than the uniform-only baseline, even though both methods solve the same downstream least-squares physics problem. The decisive difference is therefore not the corrector alone, but the geometry of the hidden basis on which that corrector operates.

For the 2D Poisson family, the guided corrector remains in the $10^{-3}$ regime across all four cases, while the best uniform-only corrector remains between $10^{-2}$ and $10^{-1}$ and deteriorates sharply for the harder extrapolative settings. The contrast is especially strong for the shifted and narrow out-of-range cases. For example, in the shifted-source case the guided corrector reaches $1.122\times10^{-3}$, whereas the best uniform-only corrector remains at $4.425\times10^{-2}$. In the narrow-source case, the gap is even larger: $1.656\times10^{-3}$ versus $3.445\times10^{-1}$. Figure~\ref{fig:ablation_poisson} shows that this gap is not closed by simply increasing the background grid resolution; the uniform-only baseline improves only gradually and remains far above the guided solution quality throughout the sweep.

The same conclusion holds, if anything more strongly, for advection--diffusion. Across all four cases, the guided corrector achieves errors between $1.846\times10^{-4}$ and $9.882\times10^{-4}$, while the best uniform-only corrector remains between $3.640\times10^{-3}$ and $3.704\times10^{-2}$. In the transport-dominated in-range case $(a,\nu)=(0.95,0.015)$, the guided corrector reaches $1.846\times10^{-4}$, whereas the best uniform-only corrector remains at $2.380\times10^{-2}$, a gap of more than two orders of magnitude. The low-viscosity out-of-range case shows the same pattern: $9.882\times10^{-4}$ for the guided corrector versus $3.704\times10^{-2}$ for the uniform-only baseline. Figure~\ref{fig:ablation_advecdiff} further shows that increasing the number of background space-time RBFs improves the uniform-only baseline only slowly and never brings it close to the guided corrector.

These ablations isolate the central mechanism of the framework. The final gain does not come merely from adding a PIELM-style least-squares solve after the predictor. Rather, it comes from using the predictor to identify \emph{where} expressive basis support is needed, and then allowing the corrector to solve the physics in that task-adaptive enriched basis. Without this guidance, the corrector is forced to distribute its capacity over a globally uniform dictionary, which is much less effective for localized elliptic responses and transport-dominated space-time structures.

\subsection{Ablation: Meta-Learned Solver vs.\ Single-Instance PINNs}
\label{subsec:ablation_single_instance}

We finally compare the KAPI predictor against a shallow single-instance PINN trained separately for one fixed PDE instance. This ablation addresses a different question from the previous uniform-grid PIELM study. There, the goal was to isolate the value of predictor-guided basis adaptation in the corrector. Here, the goal is to assess the trade-off between \emph{instance-wise optimization} and \emph{amortized parametric solving}. The single-instance PINN solves one task from scratch by directly optimizing its kernel geometry, whereas KAPI learns a shared parameter-conditioned predictor over an entire PDE family and is then evaluated on the target instance without retraining.

In both ablations, the single-instance baseline uses the same shallow Gaussian-basis representation as the corresponding KAPI predictor, but its centers, widths, gates, and coefficients are optimized specifically for one task rather than generated from a shared meta-network. We consider one representative in-range Poisson task and one representative in-range linear-advection task.

\begin{table*}[t]
	\centering
	\small
	\setlength{\tabcolsep}{4pt}
	\renewcommand{\arraystretch}{1.08}
	\begin{tabular}{llcccc}
		\toprule
		PDE & Method & Rel.\ $L^2$ error & Training time (s) & Inference time (s) & Retraining per new task \\
		\midrule
		Poisson & KAPI predictor & $1.294\times10^{-2}$ & $231.19$ & $1.843\times10^{-3}$ & No \\
		Poisson & Single-instance PINN & $5.099\times10^{-2}$ & $58.23$ & $1.132\times10^{-3}$ & Yes \\
		\midrule
		Advection & KAPI predictor & $4.022\times10^{-2}$ & $369.38$ & $4.026\times10^{-2}$ & No \\
		Advection & Single-instance PINN & $2.897\times10^{-2}$ & $150.96$ & $2.578\times10^{-2}$ & Yes \\
		\bottomrule
	\end{tabular}
	\caption{Ablation comparing the KAPI predictor against shallow single-instance PINNs on one representative Poisson task and one representative linear-advection task. KAPI incurs a larger offline training cost because it is trained over an entire parameter family, but it does not require retraining for each new PDE instance.}
	\label{tab:ablation_all}
\end{table*}

\begin{figure}[t]
	\centering
	\includegraphics[width=\textwidth]{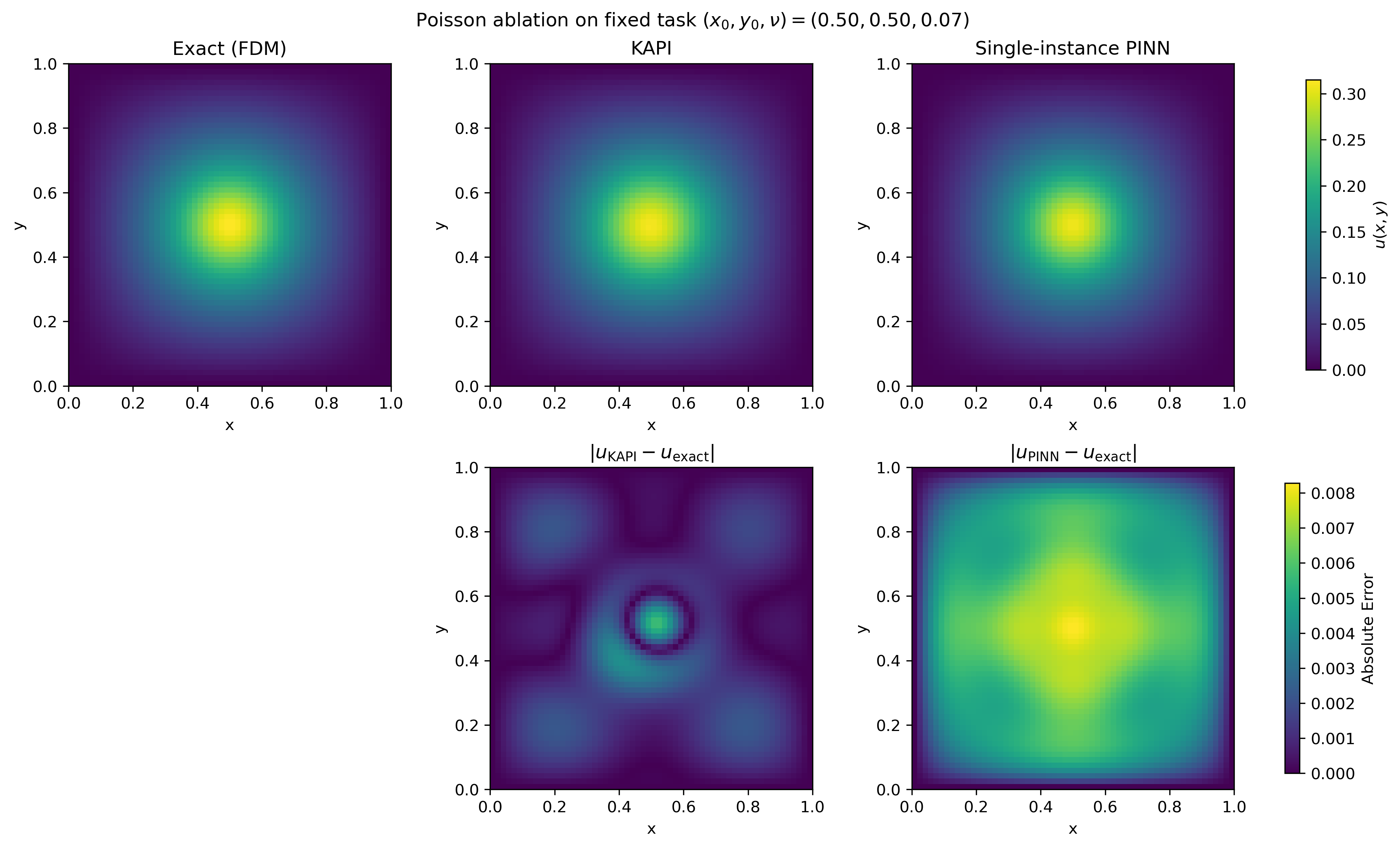}
	\caption{Poisson ablation on the fixed task $(x_0,y_0,\nu)=(0.50,0.50,0.07)$. Top row: exact finite-difference solution, KAPI prediction, and single-instance PINN prediction. Bottom row: corresponding absolute error maps. The KAPI predictor yields lower error on this representative instance, while the single-instance PINN exhibits a broader and more diffuse error pattern.}
	\label{fig:ablation_poisson_fields}
\end{figure}

\begin{figure}[t]
	\centering
	\includegraphics[width=\textwidth]{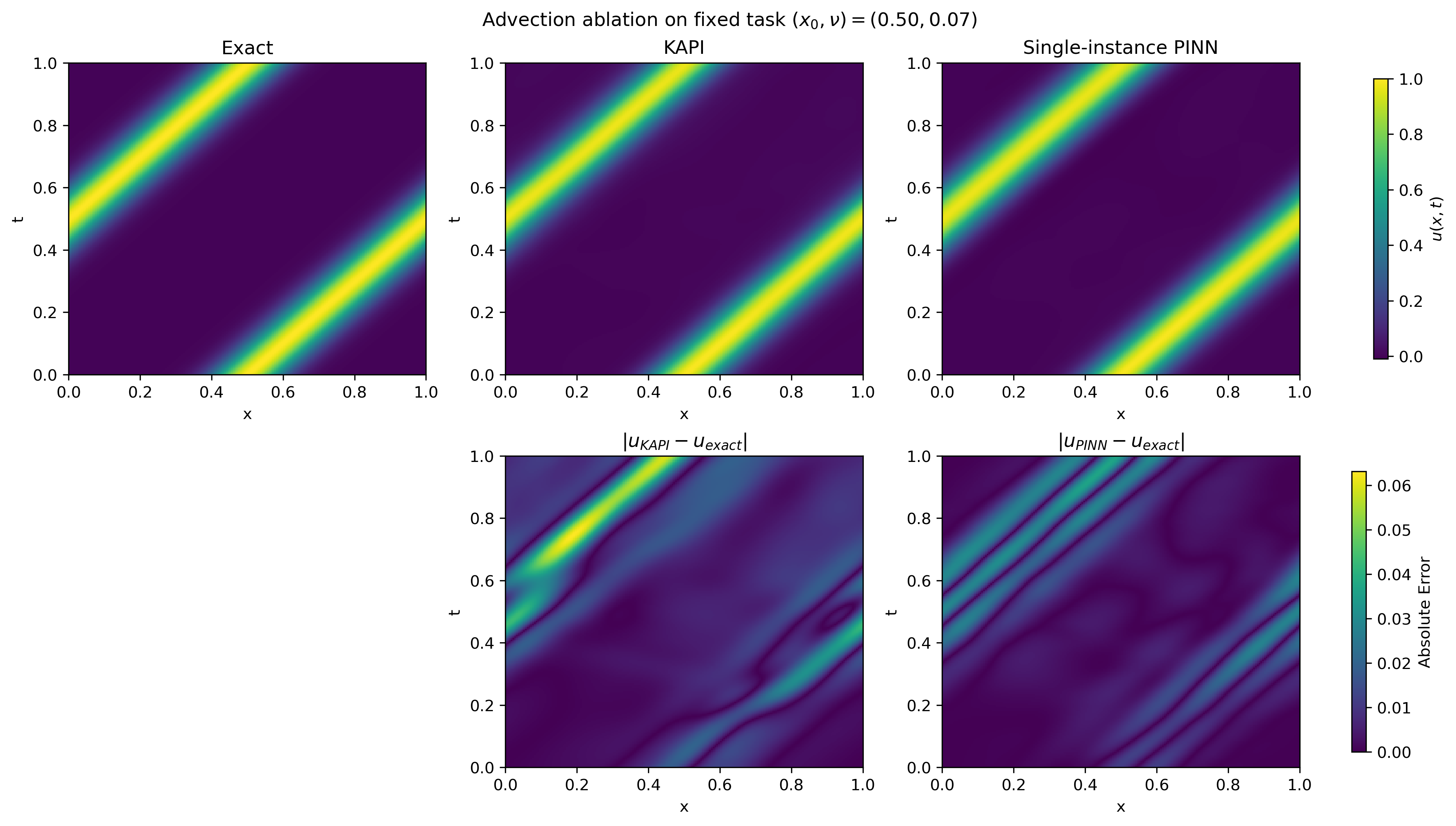}
	\caption{Advection ablation on the fixed task $(x_0,\nu)=(0.50,0.07)$. Top row: exact solution, KAPI prediction, and single-instance PINN prediction. Bottom row: corresponding absolute error maps. Both methods capture the transported ridge accurately, while the single-instance PINN is slightly sharper on this fixed instance.}
	\label{fig:ablation_advection_fields}
\end{figure}

For the Poisson task $(x_0,y_0,\nu)=(0.50,0.50,0.07)$, the KAPI predictor achieves a lower relative $L^2$ error than the single-instance PINN, improving from $5.099\times10^{-2}$ to $1.294\times10^{-2}$. Figure~\ref{fig:ablation_poisson_fields} shows that both methods recover the correct elliptic response, but the KAPI predictor produces a smaller and more localized error near the source region, whereas the single-instance PINN exhibits a broader diffuse error pattern. For the linear-advection task $(x_0,\nu)=(0.50,0.07)$, the ordering reverses: the single-instance PINN is slightly more accurate, improving from $4.022\times10^{-2}$ for KAPI to $2.897\times10^{-2}$. As seen in Fig.~\ref{fig:ablation_advection_fields}, both methods capture the transported ridge accurately, while the task-specific PINN is slightly sharper on this one fixed instance.

\begin{figure}[t]
	\centering
	\includegraphics[width=0.9\textwidth]{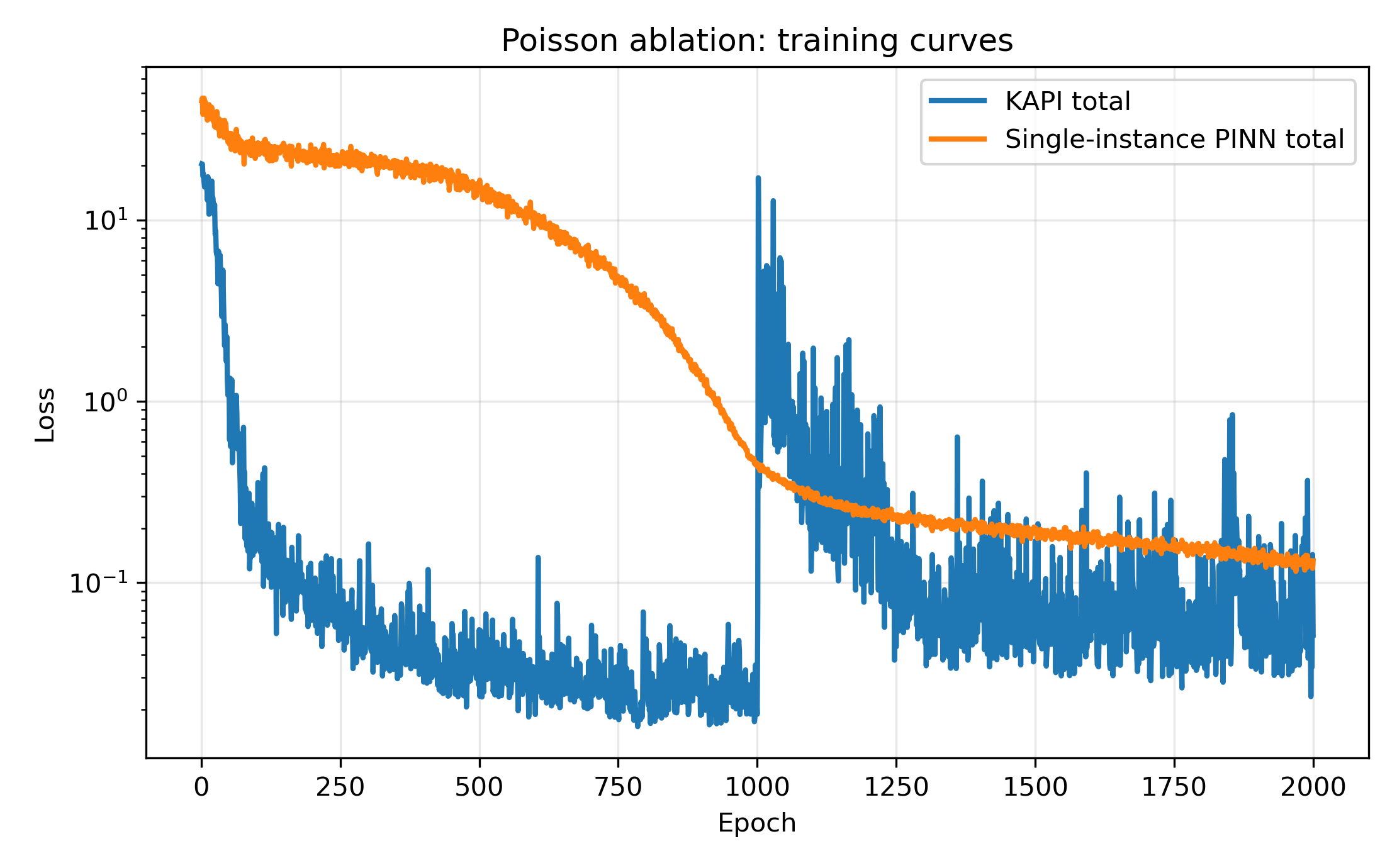}
	\caption{Training curves for the Poisson ablation. The KAPI loss is noisier because each update is computed over a batch of different tasks from the parametric family, whereas the single-instance PINN optimizes a fixed one-task objective.}
	\label{fig:ablation_poisson_training}
\end{figure}

\begin{figure}[t]
	\centering
	\includegraphics[width=0.9\textwidth]{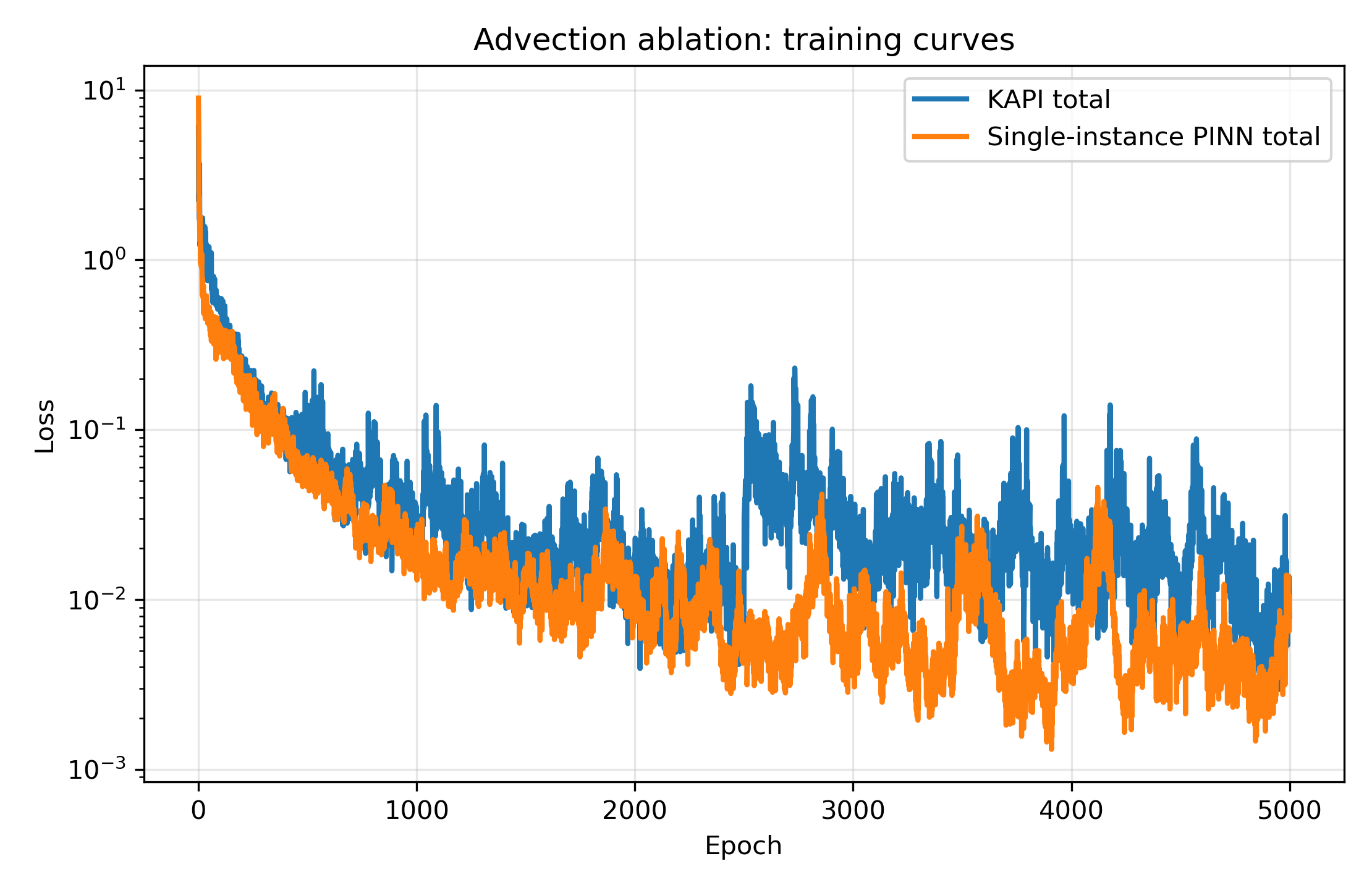}
	\caption{Training curves for the linear-advection ablation. The KAPI trajectory is noisier because each update is computed across a batch of different advection tasks, whereas the single-instance PINN optimizes only one fixed PDE objective.}
	\label{fig:ablation_advection_training}
\end{figure}

The timing results in Table~\ref{tab:ablation_all} clarify the trade-off. The single-instance PINN is cheaper to train for one task, which is expected because it optimizes only a single PDE instance. KAPI, by contrast, pays a larger offline cost to learn a reusable predictor over the full parameter family. Once trained, however, KAPI can be evaluated on new tasks without any task-specific retraining, whereas the single-instance PINN must solve a new optimization problem for every new instance. The training curves in Figs.~\ref{fig:ablation_poisson_training} and \ref{fig:ablation_advection_training} are also consistent with this interpretation: the KAPI loss is noisier because each update is computed over a batch of different tasks from the parametric family, while the single-instance PINN optimizes a fixed one-task objective.

Taken together, these ablations support a balanced conclusion. A shallow single-instance PINN can be preferable when the goal is to solve one fixed PDE instance and task-specific retraining is acceptable. KAPI, on the other hand, is designed for repeated multi-query use over a parametric family. The Poisson case shows that amortized meta-learning can even outperform a task-specific shallow PINN on a representative fixed instance, while the advection case shows the complementary situation in which task-specific optimization yields slightly better single-task accuracy. Thus, the advantage of KAPI is not that it must dominate every single-instance solver on every task, but that it provides a reusable, interpretable, and competitive family-level predictor without requiring retraining for each new PDE instance.

\section{Limitations and Future Work}
\label{sec:limitations}

The present work has several limitations. First, the current study focuses on \emph{parametric linear PDE families}. Although these test cases span markedly different regimes---including localized elliptic response, transport-dominated dynamics, mixed advection--diffusion, and variable-speed transport in both steady and unsteady settings---the predictor-corrector formulation benefits from the fact that the second-stage correction reduces to a linear least-squares problem. Extending the same framework to nonlinear PDEs will require either nonlinear correction strategies or sequential linearization schemes that preserve the efficiency and interpretability of the current approach.

Second, the method is presently developed for \emph{low-dimensional parametric families}, where the PDE is indexed by a small vector of task parameters. This setting is well suited to the goal of learning a task-to-geometry map, but it is distinct from the broader function-to-function regime targeted by neural operators. An important direction for future work is to investigate whether the same philosophy can be extended to richer conditioning information, including distributed coefficients, boundary data, or forcing fields.

Third, the current implementation uses \emph{shallow Gaussian basis models}. This design is intentional, since it makes the learned centers, widths, and activity patterns directly interpretable. However, it may also limit representational sharpness in regimes where highly anisotropic, multiscale, or strongly nonlinear structures arise. Future work could therefore explore richer kernel parameterizations, anisotropic basis families, or hybrid local--global dictionaries while retaining geometric interpretability.

Fourth, although the predictor-guided corrector is effective across a wide range of test cases, its benefit depends on the quality of the predictor-generated basis structure. This is most clearly visible in the narrow out-of-range linear-advection case, where the predictor identifies the correct transport corridor but lacks sufficient sharpness for the corrector to recover the full fine-scale structure. This suggests that future refinements may benefit from coupling the current framework with adaptive collocation, uncertainty-aware refinement, or residual-triggered local basis enrichment.

Finally, the present empirical study is limited to four canonical PDE families. While these families were chosen to span qualitatively different physical behaviors, broader validation on higher-dimensional problems, inverse settings, and more realistic scientific benchmarks would strengthen the generality of the conclusions. We view the current paper as establishing the core principle of \emph{meta-learned basis adaptation}: the next step is to test how far this principle can be pushed in more complex nonlinear and multiphysics settings.
\section{Conclusion}
\label{sec:conclusion}
We introduced a hybrid physics-informed framework for parametric linear PDEs that combines a meta-learned predictor with a least-squares corrector. The key idea is to learn, across a family of related PDEs, not the final solution directly but the \emph{geometry of the approximation space}. Given PDE parameters, the full predictor generates an interpretable task-adaptive basis through an internal meta-network, while a one-shot PIELM-style corrector solves the governing equations in this enriched basis. This yields a simple predictor-corrector architecture that combines the interpretability of shallow basis models, the flexibility of meta-learning, and the efficiency of linear least-squares physics correction.

The results show that this viewpoint is effective across diffusion-dominated, transport-dominated, mixed advection--diffusion, and variable-speed transport regimes. The predictor alone already captures meaningful physics by identifying localized source regions, transported packets, and transport-aligned space-time corridors. The corrector then exploits this geometry to obtain a substantially more accurate final solution, often improving the predictor by one or more orders of magnitude, including in several extrapolative settings. The accompanying geometry visualizations further show that the method is not behaving as a black box: the learned centers, widths, and activity patterns directly reveal how the approximation space adapts to the underlying PDE physics.

Conceptually, the proposed framework extends three important ideas from prior work. From SPINN, it inherits the view that basis geometry can be made physically meaningful and interpretable. From PIELM-style solvers, it inherits the efficiency of shallow least-squares correction. From residual-adaptive physics-informed methods, it inherits the motivation for targeted refinement. Our contribution is to lift these ideas from the single-instance setting to the \emph{parametric family} setting: the small meta-network inside the predictor acts as a family-level kernel-adaptation mechanism, replacing heuristic or iterative instance-wise basis design with an amortized task-to-geometry map.

This perspective also clarifies why the method differs from both neural operators and instance-wise adaptive solvers. Rather than learning a direct function-to-function operator map, the method learns \emph{where expressive support should be placed} for each task and then solves for the final coefficients through physics-based correction. And unlike iterative residual-adaptive refinement strategies, which are tied to one PDE instance at a time, the proposed framework amortizes basis adaptation across the full task family and deploys it in one shot at inference time after offline meta-training. The ablation studies confirm that this predictor-guided basis structure is the main source of the final gain: neither a uniform-grid PIELM corrector nor a task-specific shallow PINN achieves the same combination of accuracy, reuse, and interpretability. 

More broadly, these results suggest that, for low-dimensional parametric PDE families, learning the geometry of an approximation space may be more effective than directly learning only the solution field. This opens several directions for future work, including richer kernel parameterizations, higher-dimensional and nonlinear PDEs, and extensions that combine meta-learned basis geometry with adaptive collocation, uncertainty-aware refinement, or inverse inference while preserving the one-shot and interpretable character of the method.

\appendix
\section{Ground-Truth Computation and Reference Solutions}
\label{app:ground_truth}

\paragraph{Poisson equation.}
For the bounded-domain Poisson problem with homogeneous Dirichlet boundary conditions, no closed-form solution is available for the chosen Gaussian forcing. We therefore compute a numerical reference solution using a second-order finite-difference discretization on a uniform grid over $\Omega=[0,1]^2$, together with the standard five-point stencil for the Laplacian. The resulting linear system is solved directly and used as the reference solution in all Poisson experiments.

\paragraph{Linear advection.}
For the periodic linear advection equation, the reference solution is available in closed form:
\begin{equation}
	u(x,t;\lambda)=u_0(x-t;\lambda),
\end{equation}
with periodic wrapping on $[0,1)$. This analytic solution is used only for evaluation and is not built into either the predictor or corrector.

\paragraph{Advection--diffusion.}
For the advection--diffusion benchmark, we use the exact reference family
\begin{equation}
	u(x,t;a,\nu)=\frac{1}{\sqrt{4t+1}}
	\exp\!\left(
	-\frac{(x-x_c-a t)^2}{\nu(4t+1)}
	\right),
\end{equation}
which is also used to generate the evaluation traces and boundary data. This reference solution is used only for evaluation and is not part of the predictor or corrector hypothesis.

\paragraph{Variable-speed advection.}
For the variable-speed advection case, the reference solution is computed numerically by inverting the characteristic travel-time map associated with
\[
a(x;\beta)=1+\beta\sin(2\pi x).
\]
This reference construction is used only for evaluation and is not part of the predictor or corrector hypothesis.

\paragraph{Error metric.}
For all PDE families, we report the relative discrete $L^2$ error on a uniform evaluation grid for both the predictor and the corrected solution:
\begin{equation}
	\mathcal{E}_{\mathrm{pred}}
	=
	\frac{\|u^{\mathrm{pred}}-u_{\mathrm{ref}}\|_2}{\|u_{\mathrm{ref}}\|_2},
	\qquad
	\mathcal{E}_{\mathrm{corr}}
	=
	\frac{\|u^{\mathrm{corr}}-u_{\mathrm{ref}}\|_2}{\|u_{\mathrm{ref}}\|_2}.
\end{equation}


\section{Implementation Details}
\label{app:implementation}

This appendix summarizes the principal implementation choices used in the experiments. Since the full codebase will be released, we restrict attention to the main architectural and training settings needed to understand the reported results.

\subsection{KAPI predictor architecture}

Across all experiments, the full KAPI predictor is a shallow task-conditioned model whose basis geometry is generated by an internal parameter-conditioning network. The predictor outputs a solution value \(u^{\mathrm{pred}}(\mathbf{z};\lambda)\) for a query point \(\mathbf{z}\), while its task dependence is mediated through predictor-generated Gaussian basis parameters. These parameters include basis centers, widths, activity gates, and, for unsteady problems, dynamic amplitudes and moving centers. All predictors are trained only through physics-informed losses, without labeled solution data.

For the Poisson problem, the KAPI predictor uses \(M=128\) Gaussian kernels and an internal two-layer parameter-conditioning network of width 64. This internal network predicts spatial centers, widths, and gates, while the output coefficients are shared globally across tasks.

For periodic linear advection, the KAPI predictor uses \(M=32\) dynamic kernels. The architecture consists of an internal task encoder of width 64 together with a time-dependent network of width 96 driven by Fourier features of time. The full predictor outputs dynamic amplitudes, moving centers, and widths.

For advection--diffusion, the KAPI predictor uses \(M=48\) dynamic kernels and a three-layer temporal network of width 128 with time Fourier features. The predictor is written in residual-corrective form \(u^{\mathrm{pred}}=u_0+t(\cdot)\) so that the initial condition is built directly into the ansatz.

For variable-speed advection, the KAPI predictor uses \(M=64\) dynamic kernels and hidden width 128. The dynamic basis is generated relative to a learned periodic base dictionary, with predicted amplitudes, center shifts, and width perturbations.

\subsection{Kernel parameterization}

All KAPI predictors use Gaussian radial basis functions. In the Poisson case, the basis is defined directly in physical space through spatial centers and widths, and the homogeneous Dirichlet boundary condition is enforced exactly through the trial factor \(x(1-x)y(1-y)\).

For the one-dimensional unsteady problems, kernels are defined in space--time through dynamic centers and widths. Periodic problems use wrapped distance in space, while advection--diffusion uses the standard Euclidean distance on the bounded interval. In all unsteady cases, the basis evolves in physical space--time rather than through an explicitly prescribed characteristic coordinate.

\subsection{Predictor training procedure}

All KAPI predictors are trained with Adam using online sampling of PDE parameters and collocation points. Training begins with easier regimes and progressively expands to the full parameter range through a simple curriculum on the width or viscosity parameter.

For Poisson, we train for 2000 epochs with learning rate \(10^{-3}\), using 4 sampled tasks per batch. Interior sampling combines localized points near the source with uniformly distributed points, and a mild gate regularization is applied through optimizer weight decay.

For linear advection, we train for 5000 epochs with learning rate \(10^{-3}\) and 4 tasks per batch. Each task uses interior, initial-condition, periodic-boundary, and near-initial-time samples. The loss combines PDE, initial-condition, and periodic-boundary terms, together with mild regularization on center motion and width control.

For advection--diffusion, we train the predictor with learning rate \(10^{-3}\). Interior collocation combines uniformly sampled points with localized packet-aware samples, together with separate initial-condition and boundary samples. A viscosity curriculum is used to gradually expose the predictor to lower-viscosity regimes.

For variable-speed advection, we train for 5000 epochs with learning rate \(5\times 10^{-4}\). Interior collocation again combines uniform and localized samples, together with initial-condition and periodic-boundary samples. The predictor loss includes mild regularization on sparsity, widths, and center motion.

\subsection{Corrector construction}

For all four PDE families, the final reported solution is obtained through a predictor-guided corrector. After predictor training, the predictor-generated basis structure is frozen and converted into a PIELM-style hidden dictionary. In all cases, this dictionary is enriched by refinement and/or background basis functions to preserve global coverage and improve local resolution. The final output coefficients are then computed analytically by ridge-regularized least squares.

For the Poisson problem, the corrector does not simply reuse all predictor kernels unchanged. Instead, it selects the most informative predictor-guided spatial kernels and enriches them with additional local support around the source together with a structured background grid. The output coefficients are then fit against dense interior collocation, together with a weak anchor term derived from the predictor output. The ridge parameter is set to \(10^{-8}\).

For linear advection, the corrector uses an enriched space-time dictionary composed of selected predictor-guided dynamic centers, locally refined predictor-centered kernels, additional refinement kernels, and a modest periodic background grid. The final coefficients are selected through ridge-regularized least squares over a small grid of candidate ridge values.

For advection--diffusion, the corrector builds a frozen space-time RBF dictionary from predictor-guided inherited basis elements together with a background basis and additional refinement centers suggested by predictor-derived local structure. The final coefficients are solved by ridge-regularized least squares with regularization parameter \(10^{-8}\).

For variable-speed advection, the corrector similarly combines three sources of hidden units: a fixed background space-time basis, selected predictor-guided dynamic centers, and refinement centers extracted from predictor-informed gradient structure. The final coefficients are again obtained by ridge-regularized least squares with regularization parameter \(10^{-8}\).

\begin{table*}[t]
	\centering
	\small
	\setlength{\tabcolsep}{5pt}
	\renewcommand{\arraystretch}{1.1}
	\begin{tabular}{p{2.4cm} p{1.4cm} p{2.7cm} p{4.1cm} p{4.2cm}}
		\toprule
		\textbf{Case} & \textbf{Kernels} & \textbf{KAPI predictor} & \textbf{Main training settings} & \textbf{Corrector} \\
		\midrule
		Poisson
		& \(M=128\)
		& 2-layer internal parameter-conditioning network, width 64
		& 2000 epochs; Adam; lr \(=10^{-3}\); 4 tasks/batch; mixed localized and uniform interior sampling
		& Selected predictor-guided spatial basis with local enrichment and background support; ridge LS with weak anchor term \\
		
		Linear advection
		& \(M=32\)
		& Internal task encoder width 64; dynamic net width 96; time Fourier features
		& 5000 epochs; Adam; lr \(=10^{-3}\); 4 tasks/batch; interior + IC + periodic BC + near-IC samples
		& Selected predictor-guided spacetime basis with local refinement and periodic background; ridge LS \\
		
		Advection--diffusion
		& \(M=48\)
		& 3-layer dynamic predictor, width 128; time Fourier features
		& Adam; lr \(=10^{-3}\); uniform + localized interior samples; IC and BC samples; viscosity curriculum
		& Predictor-guided spacetime basis with refinement centers and background support; ridge LS with \(\lambda_{\mathrm{ridge}}=10^{-8}\) \\
		
		Variable-speed advection
		& \(M=64\)
		& Dynamic predictor, width 128; periodic base dictionary
		& 5000 epochs; Adam; lr \(=5\times10^{-4}\); uniform + localized interior samples; IC and periodic BC samples
		& Predictor-guided spacetime basis with gradient-informed refinement and background support; ridge LS with \(\lambda_{\mathrm{ridge}}=10^{-8}\) \\
		\bottomrule
	\end{tabular}
	\caption{Main implementation settings used across the four PDE families. We report only the principal architectural and training choices; full code and lower-level details will be released.}
	\label{tab:implementation_summary}
\end{table*}

\section{Comparison with the FiLM-HyperPINN Baseline}
\label{app:film_hyper}
In this section, we briefly summarize the hypothesis class of the proposed KAPI predictor and compare it with the FiLM-HyperPINN \cite{perez2018film} baseline used in the Poisson and advection experiments. We then describe the FiLM-HyperPINN architecture at a high level.

\subsection{Comparison of predictor hypotheses: KAPI vs. FiLM-HyperPINN}

Although both models are task-conditioned, they introduce task dependence in fundamentally different ways. FiLM-HyperPINN remains a conventional parametric PINN in which the solution is represented by a dense coordinate-based neural network, while the PDE parameters enter only through feature-wise modulation of hidden activations. By contrast, the proposed KAPI predictor is a shallow structured model in which the PDE parameters directly generate an interpretable task-dependent basis geometry.

\paragraph{Poisson problem.}
For the 2D Poisson equation with parameter vector
\[
\lambda = (x_0,y_0,\nu),
\]
the FiLM-HyperPINN baseline may be written abstractly as
\[
u_{\mathrm{FiLM}}(x,y;\lambda)
=
x(1-x)y(1-y)\,
\mathcal{N}_{\theta,\phi}\bigl((x,y);\lambda\bigr),
\]
where \(\mathcal{N}_{\theta,\phi}\) is a coordinate MLP whose hidden layers are modulated by FiLM coefficients generated from \(\lambda\). If \(h^{(0)}=(x,y)\), then the hidden layers take the form
\[
h^{(\ell+1)}
=
\tanh\!\Bigl(
\gamma^{(\ell)}(\lambda)\odot
\bigl(W^{(\ell)}h^{(\ell)}+b^{(\ell)}\bigr)
+
\beta^{(\ell)}(\lambda)
\Bigr),
\]
where \(\gamma^{(\ell)}(\lambda)\) and \(\beta^{(\ell)}(\lambda)\) are layer-wise FiLM scaling and shift vectors produced by a hypernetwork. The prefactor \(x(1-x)y(1-y)\) is used to enforce the homogeneous Dirichlet boundary condition exactly.

For the same problem, the KAPI predictor uses the shallow Gaussian basis hypothesis
\[
u_{\mathrm{KAPI}}(x,y;\lambda)
=
x(1-x)y(1-y)
\sum_{j=1}^{M}
g_j(\lambda)\,c_j\,
\exp\!\left(
-\frac{(x-\mu_{x,j}(\lambda))^2+(y-\mu_{y,j}(\lambda))^2}{\sigma_j(\lambda)^2}
\right),
\]
where the task-conditioning network maps \(\lambda\) to the gates \(g_j(\lambda)\), centers \(\mu_{x,j}(\lambda)\), \(\mu_{y,j}(\lambda)\), and widths \(\sigma_j(\lambda)\), while the coefficients \(c_j\) are shared across tasks. Thus, KAPI does not modulate a generic hidden feature space; instead, it directly generates the geometry of the approximation space itself.

\paragraph{Advection problem.}
For the 1D periodic linear advection equation with parameter vector
\[
\lambda = (x_0,\nu),
\]
the FiLM-HyperPINN baseline is written as
\[
u_{\mathrm{FiLM}}(x,t;\lambda)
=
\mathcal{N}_{\theta,\phi}\bigl(\chi(x,t);\lambda\bigr),
\]
where \(\chi(x,t)\) denotes the engineered coordinate features used as input to the trunk network, including periodic features in \(x\) and Fourier features in \(t\). The hidden layers are again modulated through FiLM:
\[
h^{(\ell+1)}
=
\tanh\!\Bigl(
\gamma^{(\ell)}(\lambda)\odot
\bigl(W^{(\ell)}h^{(\ell)}+b^{(\ell)}\bigr)
+
\beta^{(\ell)}(\lambda)
\Bigr).
\]
In this case, the model remains a dense coordinate-network ansatz, while periodic boundary conditions and the initial condition are imposed through the training loss.

In contrast, the KAPI predictor uses a dynamic periodic Gaussian basis:
\[
u_{\mathrm{KAPI}}(x,t;\lambda)
=
\sum_{j=1}^{M}
g_j(\lambda)\,\alpha_j(t,\lambda)\,
\exp\!\left(
-\frac{d_{\mathrm{per}}(x,\xi_j(t,\lambda))^2}{2\,h_j(t,\lambda)^2}
\right),
\]
where \(d_{\mathrm{per}}\) denotes the wrapped periodic distance. Here the task-conditioning mechanism first determines the task-adaptive basis configuration, and a time-dependent network then generates the amplitudes \(\alpha_j(t,\lambda)\), moving centers \(\xi_j(t,\lambda)\), and widths \(h_j(t,\lambda)\). Hence, the advection-version of KAPI is again a basis-generating model rather than a feature-modulated coordinate MLP.

\paragraph{Main architectural distinction.}
The essential distinction is therefore the following. FiLM-HyperPINN adapts a black-box deep neural representation through layer-wise feature modulation, whereas KAPI constructs an explicit, shallow, and interpretable task-dependent basis. In KAPI, the PDE parameters determine where basis functions are placed, how wide they are, and which ones are active. This basis-generating structure is central to the proposed predictor-corrector framework, because the predictor-generated kernels can be passed directly to the second-stage least-squares corrector. Such a transfer is not naturally available in the FiLM-HyperPINN architecture.

\subsection{FiLM-HyperPINN baseline architecture}

The FiLM-HyperPINN baseline used in this work follows a simple two-part design inspired by feature-wise linear modulation (FiLM) \cite{perez2018film}. The first part is a \emph{trunk network}, which maps the physical coordinates to a latent representation. The second part is a small \emph{hypernetwork}, which takes the PDE parameters as input and outputs layer-wise FiLM coefficients \((\gamma,\beta)\) used to affine-modulate the hidden activations of the trunk. In this way, a single PINN is conditioned to represent an entire parametric PDE family.

For the Poisson case, the trunk network takes the spatial coordinate \((x,y)\) as input and is implemented as a fully connected multilayer perceptron with four hidden layers of width \(128\), using \(\tanh\) activations. The conditioning vector is \(\lambda=(x_0,y_0,\nu)\). A separate hypernetwork with two hidden layers of width \(64\) maps \(\lambda\) to the FiLM coefficients for all hidden layers of the trunk. The final scalar output is multiplied by the factor \(x(1-x)y(1-y)\), so that the homogeneous Dirichlet boundary condition is satisfied exactly. Training is fully physics-informed through the Poisson residual, without using paired solution snapshots.

For the advection case, the trunk network again has four hidden layers of width \(128\) with \(\tanh\) activations, but its input is augmented with engineered coordinate features tailored to periodic transport, namely periodic spatial features and Fourier features in time. The conditioning vector is \(\lambda=(x_0,\nu)\), corresponding to the center and width of the Gaussian initial packet. The hypernetwork again uses two hidden layers of width \(64\) to generate the FiLM coefficients that modulate the trunk. Unlike the Poisson case, no hard output transform is imposed; instead, the model is trained using a physics-informed loss containing the advection residual together with initial-condition and periodic-boundary losses.

Overall, FiLM-HyperPINN provides a compact parametric PINN baseline that shares a common trunk across the PDE family while adapting hidden representations through FiLM modulation. Its role in the present work is to serve as a representative task-conditioned deep neural baseline against which the structured basis-generating KAPI predictor can be compared.

\section{Comparison with the Physics-Informed DeepONet Baseline}
\label{app:pideeponet}
In this section, we summarize the hypothesis class of the proposed KAPI predictor and compare it with the physics-informed DeepONet baseline used in the Poisson and advection experiments. We then briefly describe the PI-DeepONet architecture used in these comparisons. For implementation, please refer to \url{https://github.com/PredictiveIntelligenceLab/Physics-informed-DeepONets/blob/main/Advection/PI_DeepONet_adv.ipynb}. 

\subsection{Comparison of predictor hypotheses: KAPI vs. physics-informed DeepONet}

Although both models are trained through physics-informed losses and are designed to handle parametric PDE families, they represent task dependence in very different ways. The physics-informed DeepONet baseline follows the standard branch--trunk philosophy: the PDE instance is first encoded through sampled sensor values, and the solution is then produced by coupling this task encoding with coordinate-dependent trunk features. By contrast, the proposed KAPI predictor is a shallow task-conditioned basis model in which the PDE parameters directly generate the approximation-space geometry itself.

\paragraph{Poisson problem.}
For the 2D Poisson problem with parameter vector
\[
\lambda = (x_0,y_0,\nu),
\]
the physics-informed DeepONet baseline may be written as
\[
u_{\mathrm{PI\mbox{-}DeepONet}}(x,y;\lambda)
=
x(1-x)y(1-y)
\left(
\sum_{k=1}^{L}
b_k(v_\lambda)\,t_k(x,y)
+
b_0
\right),
\]
where \(v_\lambda\) denotes the branch input associated with the task \(\lambda\), constructed here by sampling the Gaussian source term at a fixed set of sensor locations. The branch network maps \(v_\lambda\) to latent coefficients
\[
b(v_\lambda) = \bigl(b_1(v_\lambda),\dots,b_L(v_\lambda)\bigr),
\]
while the trunk network maps the query coordinate \((x,y)\) to latent trunk features
\[
t(x,y) = \bigl(t_1(x,y),\dots,t_L(x,y)\bigr).
\]
The final prediction is obtained by their latent inner product, followed by multiplication with the boundary factor \(x(1-x)y(1-y)\) so that the homogeneous Dirichlet boundary condition is satisfied exactly.

For the same problem, the KAPI predictor uses the shallow Gaussian basis hypothesis
\[
u_{\mathrm{KAPI}}(x,y;\lambda)
=
x(1-x)y(1-y)
\sum_{j=1}^{M}
g_j(\lambda)\,c_j\,
\exp\!\left(
-\frac{(x-\mu_{x,j}(\lambda))^2+(y-\mu_{y,j}(\lambda))^2}{\sigma_j(\lambda)^2}
\right).
\]
Here the task-conditioning map acts directly on the basis geometry through the gates \(g_j(\lambda)\), centers \(\mu_{x,j}(\lambda)\), \(\mu_{y,j}(\lambda)\), and widths \(\sigma_j(\lambda)\), whereas the coefficients \(c_j\) are shared across tasks. Thus, KAPI directly generates a task-adaptive approximation space, while PI-DeepONet instead generates a latent operator representation through branch and trunk embeddings.

\paragraph{Advection problem.}
For the 1D periodic linear advection problem with parameter vector
\[
\lambda = (x_0,\nu),
\]
the physics-informed DeepONet baseline is built from a sensor representation of the initial condition. Its hypothesis may be written as
\[
u_{\mathrm{PI\mbox{-}DeepONet}}(x,t;\lambda)
=
u_{0,\mathrm{rec}}(x;v_\lambda)
+
t\,\mathcal{D}\!\left(
\tau(x,t),\,b(v_\lambda),\,\tau(x,t)\odot b(v_\lambda)
\right),
\]
where \(v_\lambda\) denotes the branch input obtained by sampling the periodic Gaussian initial condition at fixed sensor points, \(\tau(x,t)\) is the trunk feature vector computed from engineered periodic/Fourier features of \((x,t)\), and \(\mathcal{D}\) is a small decoder network applied after branch--trunk fusion. The first term \(u_{0,\mathrm{rec}}(x;v_\lambda)\) is a separate branch--trunk reconstruction of the initial condition, and the factor \(t\) hard-enforces this anchor at \(t=0\). Thus, the advection PI-DeepONet remains a latent branch--trunk model, but with a hard initial-condition decomposition.

By contrast, the KAPI predictor for advection uses a dynamic periodic Gaussian basis:
\[
u_{\mathrm{KAPI}}(x,t;\lambda)
=
\sum_{j=1}^{M}
g_j(\lambda)\,\alpha_j(t,\lambda)\,
\exp\!\left(
-\frac{d_{\mathrm{per}}(x,\xi_j(t,\lambda))^2}{2\,h_j(t,\lambda)^2}
\right),
\]
where \(d_{\mathrm{per}}\) denotes the wrapped periodic distance. In this case, the task-conditioning mechanism determines the active basis geometry, and a time-dependent network generates the amplitudes \(\alpha_j(t,\lambda)\), moving centers \(\xi_j(t,\lambda)\), and widths \(h_j(t,\lambda)\). Hence, KAPI again remains an explicit basis-generating model rather than a branch--trunk latent operator network.

\paragraph{Main architectural distinction.}
The essential architectural distinction is therefore the following. Physics-informed DeepONet represents a task through sensor-sampled function values and combines this task encoding with coordinate-dependent trunk features in a learned latent space. In contrast, KAPI maps the PDE parameters directly to an explicit task-dependent basis geometry. Thus, PI-DeepONet learns a black-box branch--trunk operator representation, whereas KAPI learns an interpretable approximation space whose centers, widths, and activity patterns are directly tied to the physics of the PDE family. This distinction is especially important in the present work because the KAPI predictor produces basis geometry that can be transferred directly to the second-stage least-squares corrector, while PI-DeepONet does not expose such a transferable basis structure.

\subsection{Physics-informed DeepONet baseline architecture}

The physics-informed DeepONet baseline used in this work follows a standard branch--trunk design, with problem-specific modifications for the Poisson and advection cases. In both settings, the \emph{branch network} receives a sensor-based representation of the PDE instance, while the \emph{trunk network} receives the physical query coordinate. The model is trained by minimizing physics-informed residual losses, rather than by supervised regression on full solution fields.

For the Poisson case, the branch input is formed by evaluating the Gaussian source term associated with \(\lambda=(x_0,y_0,\nu)\) at a fixed set of sensor locations in the spatial domain. This sensor vector is passed through a branch MLP, while the coordinate \((x,y)\) is passed through a trunk MLP. Both branch and trunk networks use fully connected layers with \(\tanh\) activations and produce latent vectors of the same dimension. The scalar output is obtained as an inner product between the branch and trunk latent vectors, plus a learnable bias. The final prediction is multiplied by the factor \(x(1-x)y(1-y)\), thereby enforcing the homogeneous Dirichlet boundary condition exactly. In this way, the Poisson PI-DeepONet baseline serves as a source-to-solution operator learner trained through the Poisson residual.

For the advection case, the branch input is formed by sampling the periodic Gaussian initial condition corresponding to \(\lambda=(x_0,\nu)\) at a fixed set of periodic sensor points. The architecture is slightly richer than in the Poisson case. A feature transform first maps the query coordinate \((x,t)\) to engineered features consisting of periodic Fourier features in space and multiscale Fourier/polynomial features in time. These transformed coordinates are passed through the trunk network, while the sensor vector is passed through the branch network. Their interaction is then fused through concatenation of trunk features, branch features, and their elementwise product, followed by a decoder MLP. In addition, a separate branch--trunk pair reconstructs the initial condition, and the final prediction is written as
\[
u(x,t) = u_{0,\mathrm{rec}}(x) + t\,u_{\mathrm{corr}}(x,t),
\]
so that the initial condition is enforced exactly at \(t=0\). Periodicity is represented through the spatial Fourier features and through the periodic construction of the branch input.

Overall, the physics-informed DeepONet baseline provides a representative neural-operator-style parametric baseline. Its role in the present work is to test whether a learned branch--trunk latent representation of the PDE family can compete with the structured, interpretable, basis-generating KAPI predictor.

\section{Multi-Source Poisson Extensions}
\label{app:poisson_extend}
To assess whether the predictor-guided basis adaptation extends beyond the single-source Poisson family used in the main text, we also considered Poisson test families with two and four Gaussian forcing terms of common width and separated source centers. In these extensions, the task parameter vector was enlarged to encode all source locations together with the common width parameter, while the same two-stage KAPI workflow was retained: a predictor was first meta-trained to generate task-adaptive basis geometry, and a second-stage corrector then inherited this geometry and solved for the final coefficients by least squares.

For the two-source case, the predictor successfully identified both active response regions, and the corrector further reduced the error substantially across representative test cases. For the four-source case, despite the more complex multi-peak forcing geometry, the same methodology remained effective after increasing the predictor and corrector capacity appropriately. These results indicate that the KAPI framework is not restricted to single-source localization, but extends naturally to multi-modal forcing patterns with multiple separated regions of activity.

\begin{figure*}[t]
	\centering
	\includegraphics[width=\textwidth]{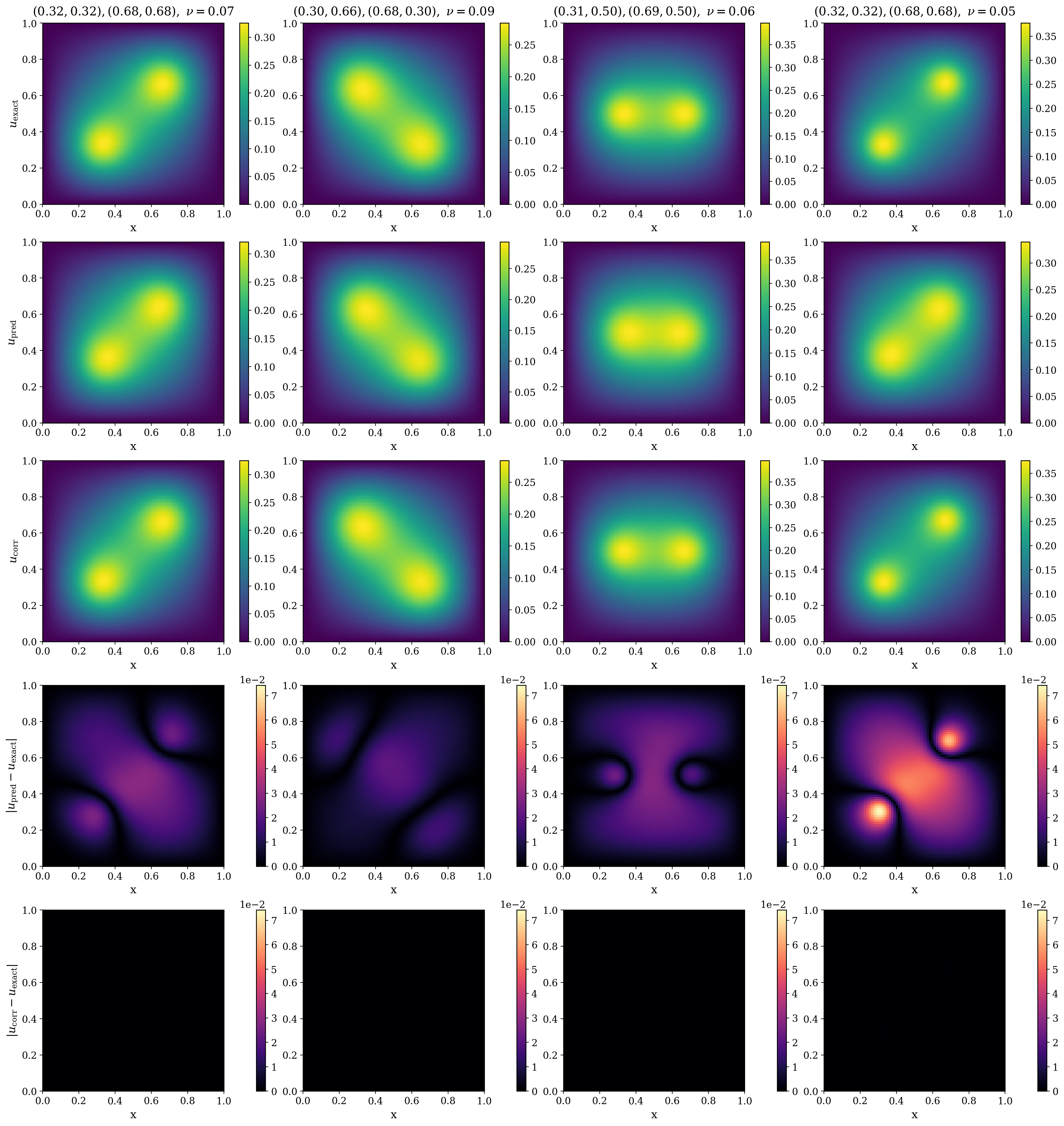}
	\caption{
		Two-source extension of the 2D Poisson predictor-corrector experiment. Each column shows one test case; rows show the reference solution, predictor, corrected solution, predictor error, and corrector error.
	}
	\label{fig:poisson_two_source_extension}
\end{figure*}

\begin{figure*}[t]
	\centering
	\includegraphics[width=\textwidth]{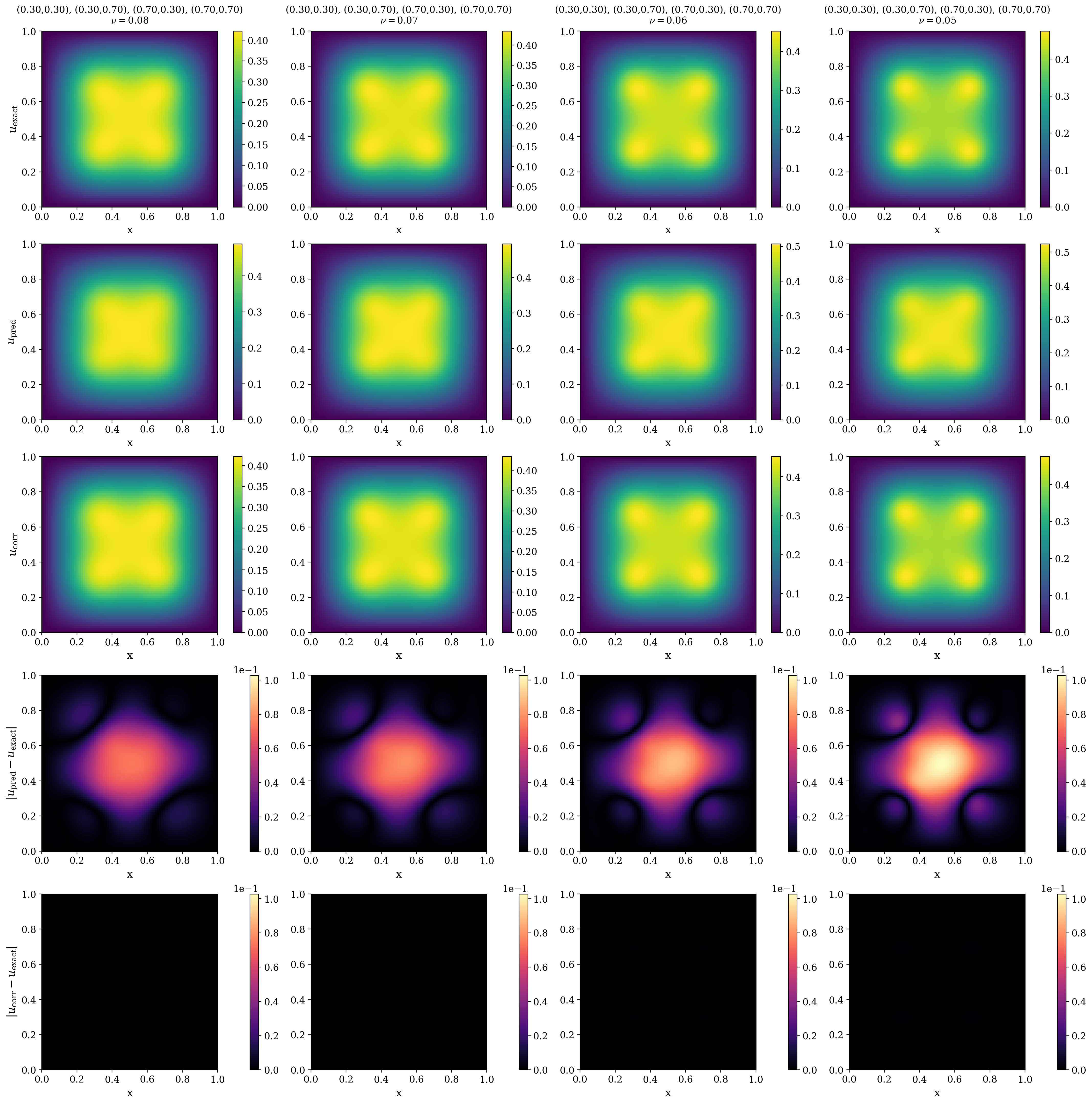}
	\caption{
		Four-source extension of the 2D Poisson predictor-corrector experiment. Each column shows one test case; rows show the reference solution, predictor, corrected solution, predictor error, and corrector error.
	}
	\label{fig:poisson_four_source_extension}
\end{figure*}

\begin{figure}[t]
	\centering
	\begin{subfigure}[b]{0.48\textwidth}
		\centering
		\includegraphics[width=\textwidth]{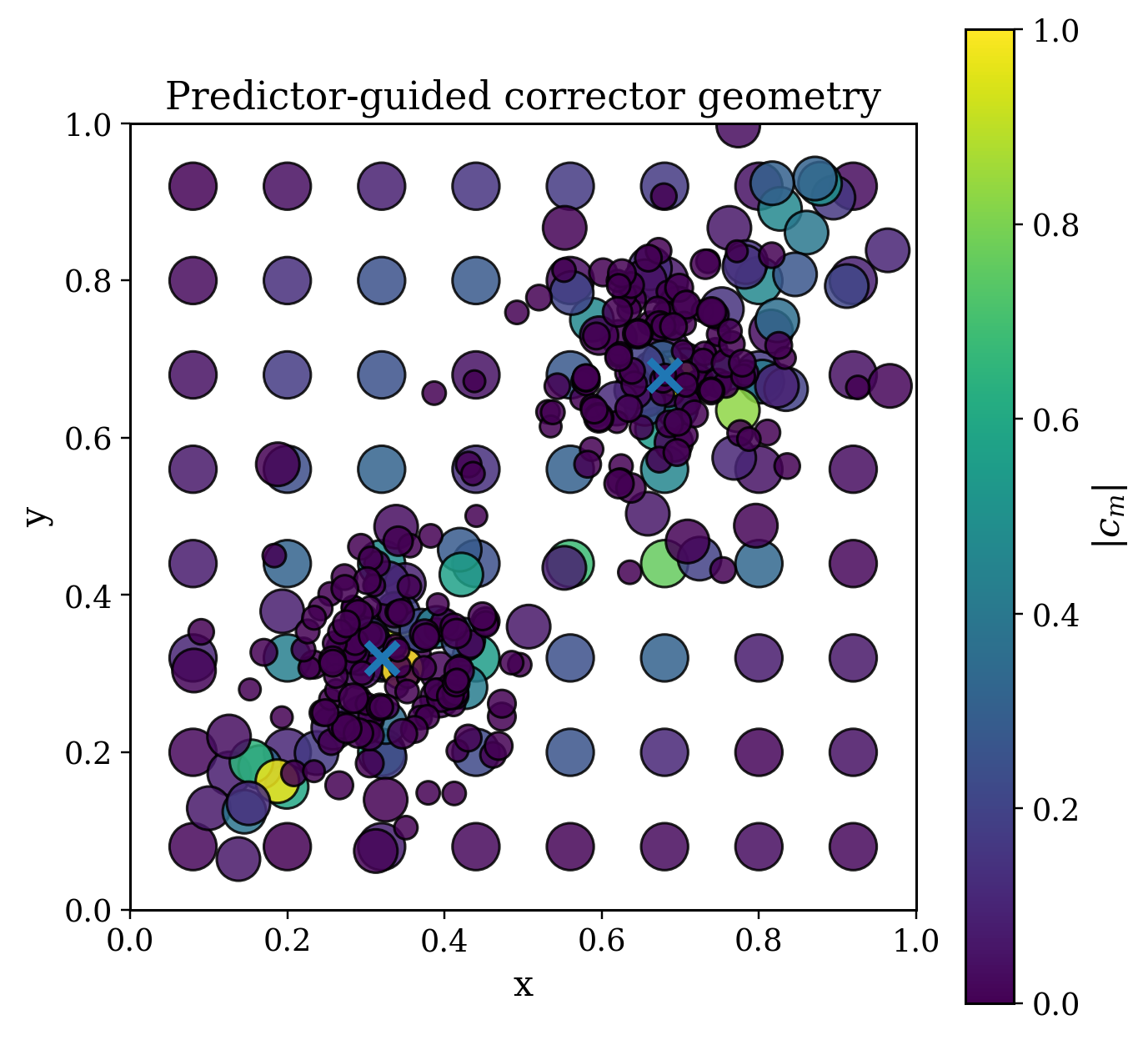}
		\caption{Two-source case.}
		\label{fig:two_source_geometry}
	\end{subfigure}
	\hfill
	\begin{subfigure}[b]{0.48\textwidth}
		\centering
		\includegraphics[width=\textwidth]{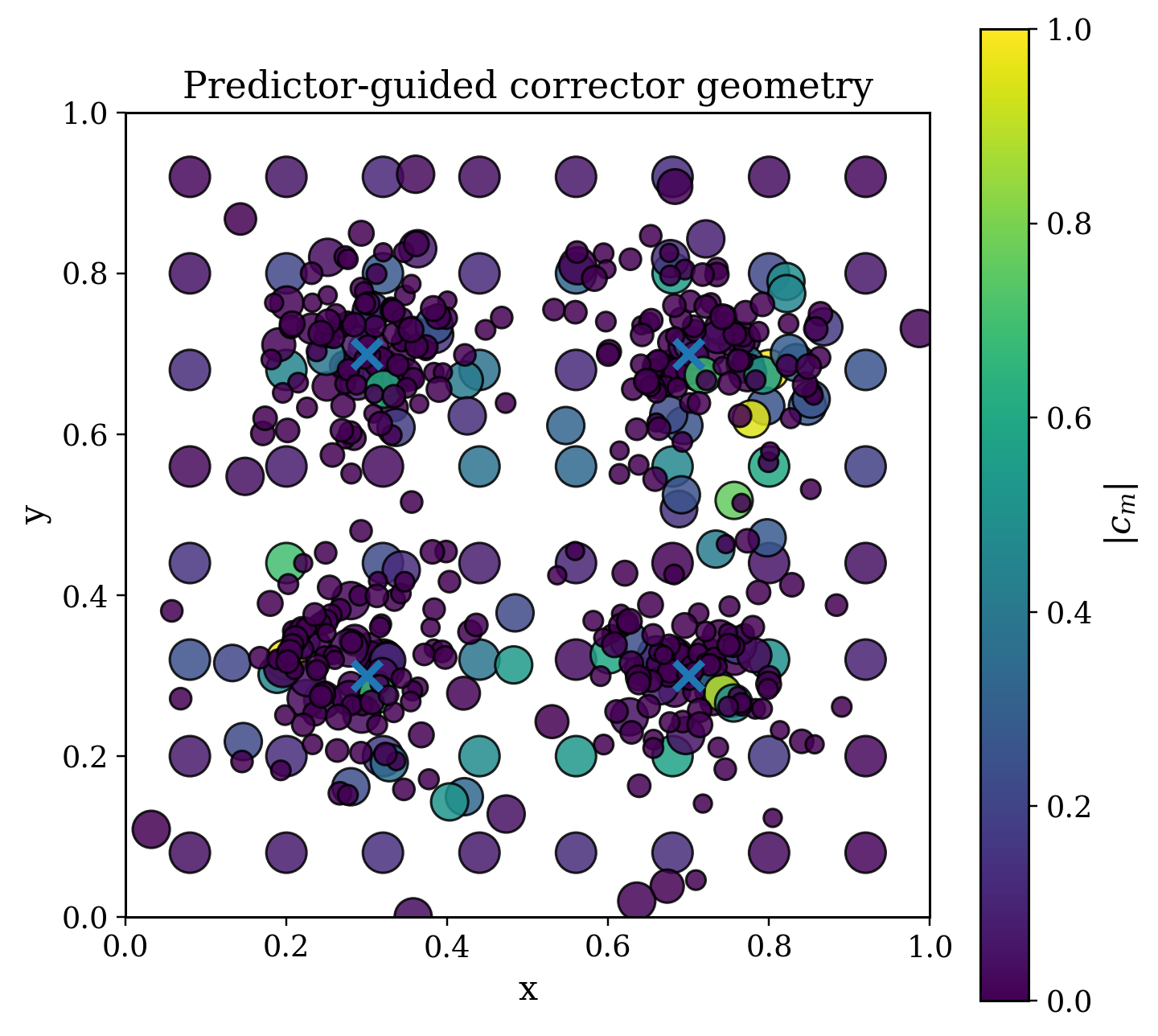}
		\caption{Four-source case.}
		\label{fig:four_source_geometry}
	\end{subfigure}
	\caption{
		Predictor-guided corrector geometries for the multi-source Poisson extensions.
	}
	\label{fig:multisource_geometry}
\end{figure}

\section{Advection with a Mexican-hat initial condition}
\label{app:mexhat_advection}

\begin{figure}[t]
	\centering
	\includegraphics[width=\textwidth]{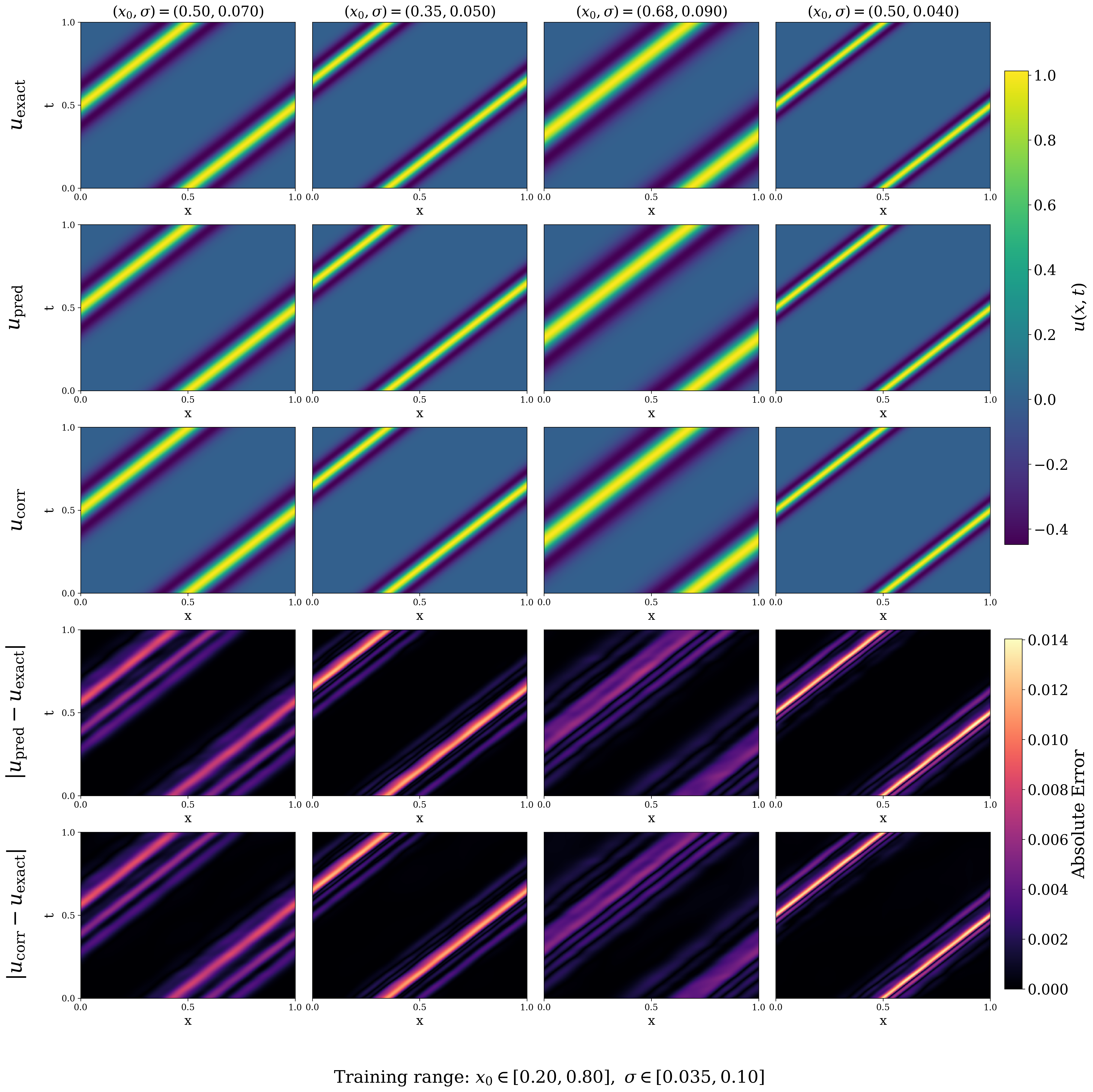}
	\caption{
		Predictor-corrector performance for the Mexican-hat advection test. Rows show \(u_{\mathrm{exact}}\), \(u_{\mathrm{pred}}\), \(u_{\mathrm{corr}}\), \(|u_{\mathrm{pred}}-u_{\mathrm{exact}}|\), and \(|u_{\mathrm{corr}}-u_{\mathrm{exact}}|\).
	}
	\label{fig:mexhat_advection}
\end{figure}

\begin{figure}[t]
	\centering
	\includegraphics[width=\textwidth]{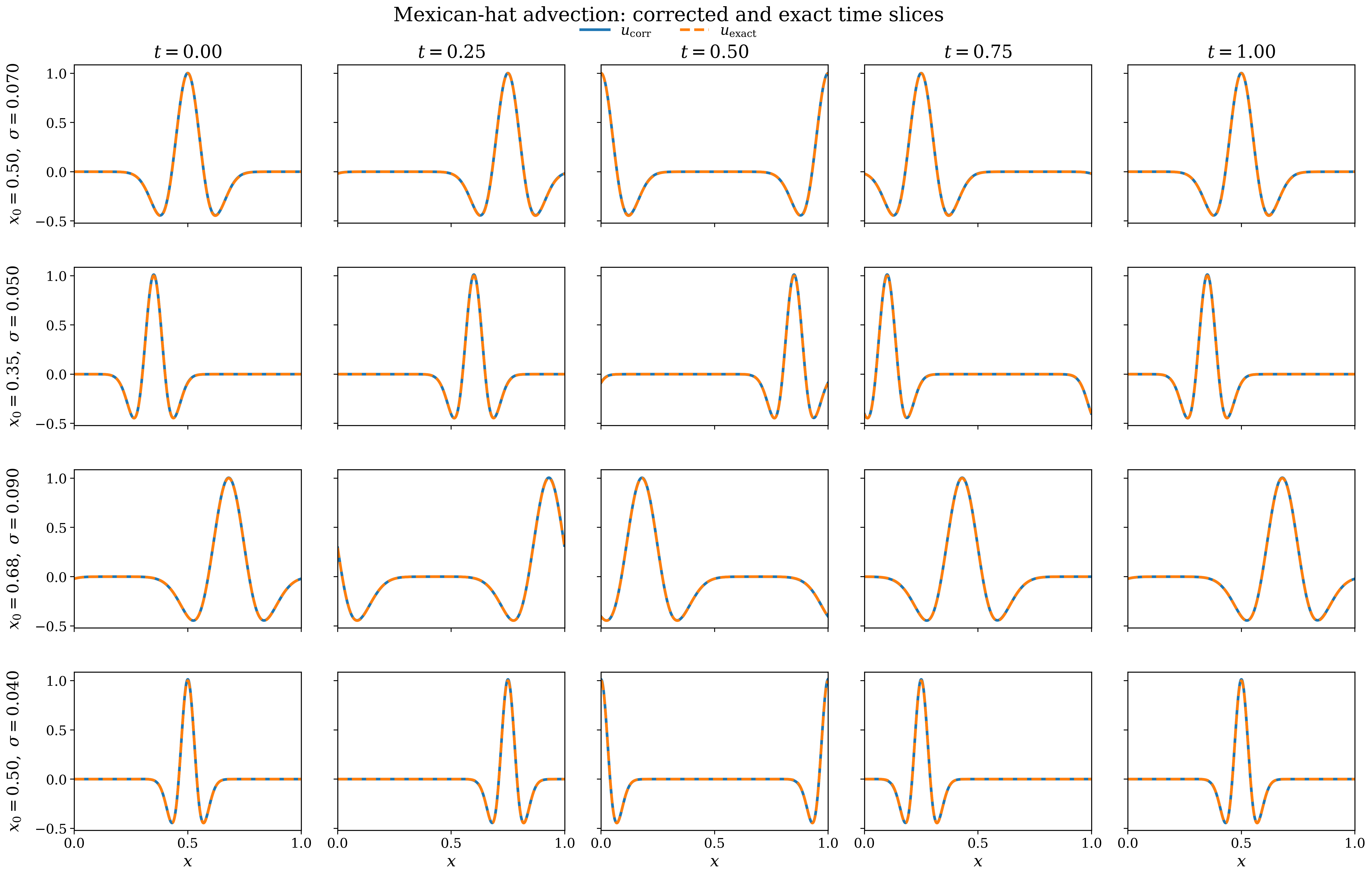}
	\caption{
		Time-slice comparison for the Mexican-hat advection test. Solid blue curves show \(u_{\mathrm{corr}}\) and dashed red curves show \(u_{\mathrm{exact}}\) at \(t=0,0.25,0.5,0.75,1.0\).
	}
	\label{fig:mexhat_advection_slices}
\end{figure}

To test whether the advection predictor-corrector is tied to Gaussian initial data, we also considered the 1D periodic linear advection problem with a periodicized Mexican-hat initial condition
\[
u_t + u_x = 0, \qquad x\in[0,1), \quad t\in[0,1],
\]
\[
u(x,0)=u_0(x;x_0,\sigma)
=
\sum_{k=-1}^{1}
\left(
1-\frac{(x-x_0+k)^2}{\sigma^2}
\right)
\exp\!\left(
-\frac{(x-x_0+k)^2}{2\sigma^2}
\right).
\]
Here \(x_0\) denotes the packet center and \(\sigma\) its width. The predictor was trained over
\[
x_0\in[0.20,0.80], \qquad \sigma\in[0.035,0.10],
\]
and the same Stage-2 advection corrector pipeline as in the Gaussian case was retained. Representative spacetime results are shown in Fig.~\ref{fig:mexhat_advection}, while time-slice comparisons are shown in Fig.~\ref{fig:mexhat_advection_slices}. Quantitative errors are reported in Table~\ref{tab:mexhat_advection}. The predictor remains highly accurate on this non-Gaussian, sign-changing profile, and the corrector provides small but consistent improvements without degrading the solution.

\begin{table}[t]
	\centering
	\caption{Relative \(L^2\) errors for the Mexican-hat advection test cases.}
	\label{tab:mexhat_advection}
	\small
	\setlength{\tabcolsep}{5pt}
	\renewcommand{\arraystretch}{1.08}
	\begin{tabular}{lcc}
		\toprule
		Test case \((x_0,\sigma)\) & \(\mathcal{E}_{\mathrm{pred}}\) & \(\mathcal{E}_{\mathrm{corr}}\) \\
		\midrule
		\((0.50,\,0.070)\) & \(8.541\times10^{-3}\) & \(8.398\times10^{-3}\) \\
		\((0.35,\,0.050)\) & \(1.039\times10^{-2}\) & \(1.027\times10^{-2}\) \\
		\((0.68,\,0.090)\) & \(5.903\times10^{-3}\) & \(5.677\times10^{-3}\) \\
		\((0.50,\,0.040)\) & \(1.105\times10^{-2}\) & \(1.100\times10^{-2}\) \\
		\bottomrule
	\end{tabular}
\end{table}

\section*{CRediT authorship contribution statement}
\textbf{Vikas Dwivedi:} Conceptualization, Methodology, Software, Writing - Original Draft, 
\textbf{Bruno Sixou} and \textbf{Monica Sigovan:} Supervision and Writing - Review \& Editing.

\section*{Acknowledgements}
This work was supported by the ANR (Agence Nationale de la Recherche), France, through the RAPIDFLOW project (Grant no. ANR-24-CE19-1349-01). 

\bibliographystyle{unsrt}  
\bibliography{references}

\end{document}